\documentclass[10pt,journal,compsoc]{IEEEtran}

\usepackage{booktabs} 
\usepackage{tabu}

\usepackage{color}

\usepackage{bookmark}
\usepackage{multirow}
\usepackage{array}
\usepackage[normalem]{ulem}

\makeatletter  
\newif\if@restonecol  
\makeatother

\usepackage[linesnumbered,ruled,vlined]{algorithm2e} 
\usepackage{algpseudocode}
\usepackage{amsmath}  
\usepackage{amssymb}

\usepackage{ragged2e}

\usepackage{cite}
\usepackage[pdftex]{graphicx}
  
\graphicspath{IMAGES/}
\DeclareGraphicsExtensions{.pdf,.jpeg,.png}

\usepackage{threeparttable}
\usepackage{microtype}

\usepackage{dblfloatfix}
\usepackage{diagbox}

\usepackage{url}

\usepackage{lipsum}
\usepackage{xcolor}
\usepackage{listings}

\begin{document}

\title{M-Evolve: Structural-Mapping-Based Data Augmentation for Graph Classification}

\author{Jiajun Zhou,
		Jie Shen,
		Shanqing~Yu,
		Guanrong Chen,~\IEEEmembership{Fellow,~IEEE},
		and~Qi~Xuan,~\IEEEmembership{Member,~IEEE}
\IEEEcompsocitemizethanks{
\IEEEcompsocthanksitem J. Zhou, J. Shen, S. Yu and Q. Xuan are with the Institute of Cyberspace Security, College of Information Engineering, Zhejiang University of Technology, Hangzhou 310023, China. E-mail: \{jjzhou, shenj, yushanqing, xuanqi\}@zjut.edu.cn).
\IEEEcompsocthanksitem G. Chen is with the Department of Electrical Engineering, City University of Hong Kong, Hong Kong SAR, China. E-mail: eegchen@cityu.edu.hk.
\IEEEcompsocthanksitem Corresponding author: Qi Xuan.
}

}

\markboth{IEEE TRANSACTIONS ON NETWORK SCIENCE AND ENGINEERING}%
{Shell \MakeLowercase{\textitit{et al.}}: Bare Demo of IEEEtran.cls for Computer Society Journals}


\IEEEtitleabstractindextext{%
\justifying  
\begin{abstract}
	Graph classification, which aims to identify the category labels of graphs, plays a significant role in drug classification, toxicity detection, protein analysis etc.
	However, the limitation of scale in the benchmark datasets makes it easy for graph classification models to fall into over-fitting and undergeneralization.
	To improve this, we introduce data augmentation on graphs (i.e. graph augmentation) and present four methods: {\em{random mapping}}, {\em{vertex-similarity mapping}}, {\em{motif-random mapping}} and {\em{motif-similarity mapping}}, 
	to generate more weakly labeled data for small-scale benchmark datasets via heuristic transformation of graph structures. 
	Furthermore, we propose a generic model evolution framework, named \textit{M-Evolve}, which combines graph augmentation, data filtration and model retraining to optimize pre-trained graph classifiers.
	Experiments on six benchmark datasets demonstrate that the proposed framework helps existing graph classification models alleviate over-fitting and undergeneralization in the training on small-scale benchmark datasets, which successfully yields an average improvement of 3 – 13\% accuracy on graph classification tasks.
\end{abstract}

\begin{IEEEkeywords}
	Graph classification; Data augmentation; Model evolution; Vertex similarity; Motif.
\end{IEEEkeywords}}

\maketitle

\IEEEdisplaynontitleabstractindextext

\IEEEpeerreviewmaketitle

\IEEEraisesectionheading{\section{Introduction}\label{sec:introduction}}
\IEEEPARstart{G}{raph} 
classification, or network classification, which aims to identify the category labels of graphs in a dataset, has recently attracted considerable attention from different fields like bioinformatics~\cite{borgwardt2005protein}, chemoinformatics~\cite{duvenaud2015convolutional} and social network~\cite{yanardag2015deep}.
For instance, in bioinformatics, protein or enzymes can be represented as labeled graphs, in which vertices are atoms and edges represent chemical bonds that connect atoms.
The task of graph classification is to classify these molecular graphs according to their chemical properties like carcinogenicity, mutagenicity and toxicity.

However, in bioinformatics and chemoinformatics, the scale of the known benchmark graph datasets is generally in the range of tens to thousands, which is far from that of the available real-world bibliography datasets like COLLAB and IMDB~\cite{yanardag2015deep}.
Despite the advances of various graph classification methods, from graph kernels, graph embedding to graph neural networks, the limitation of data scale makes them easily fall into the dilemma of over-fitting and undergeneralization.
Over-fitting refers to a modeling error that occurs when a model learns a function with high variance to perfectly fit the limited set of data. 
A natural idea to address over-fitting at the data level is data augmentation, which is widely applied in computer vision.
Data augmentation encompasses a number of techniques that enhance both the scale and the quality of training data such that the models of higher performance can be learnt satisfactorily. In computer vision, image augmentation methods include geometric transformation, color depth adjustment, neural style transfer, adversarial training, etc.
\begin{figure}[htb]
	\centering
	\includegraphics[width=\linewidth]{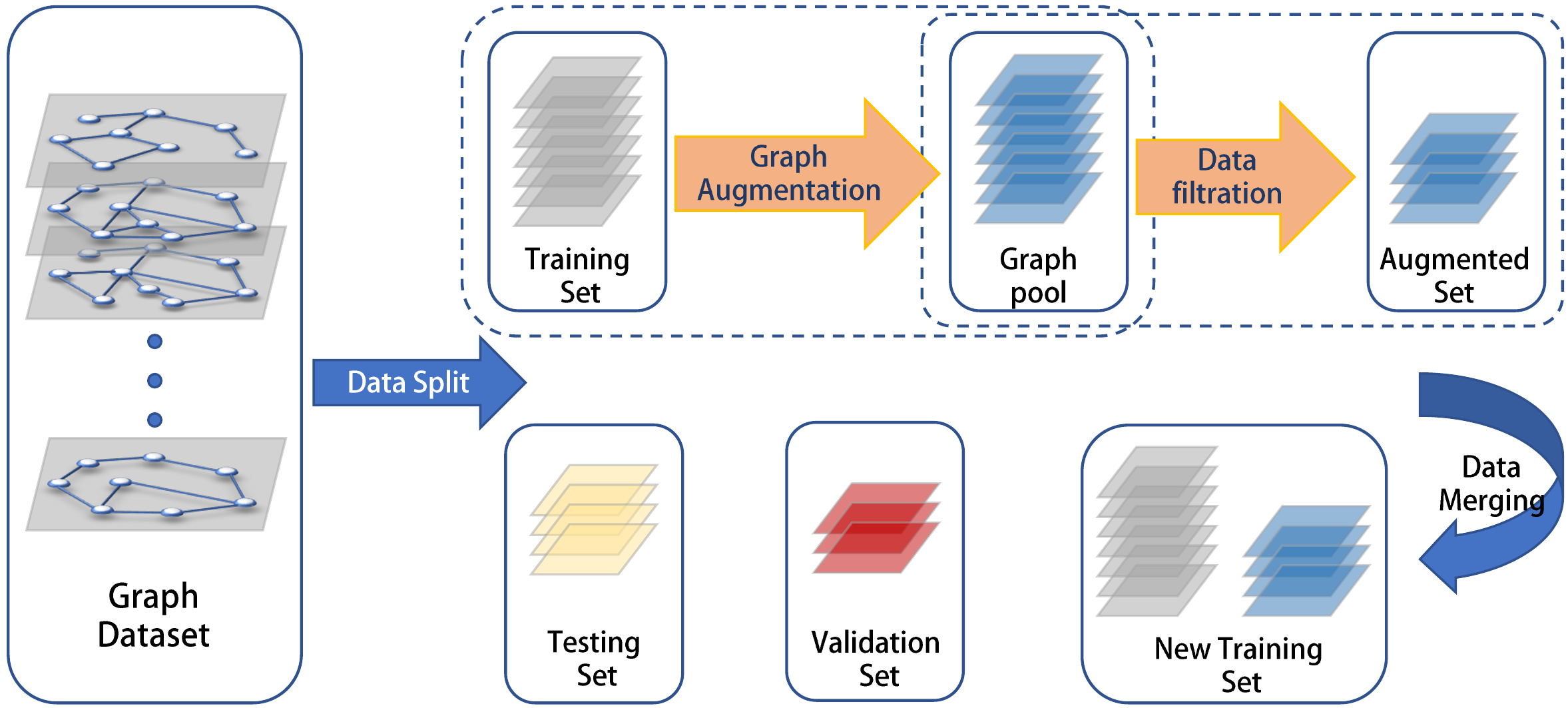}
	\caption{An illustration of data augmentation application in graph (Graph Augmentation).}
	\label{fig:framework}
  \end{figure}
However, different from image data, which have a clear grid structure, graphs have irregular topological structures, making it hard to generalize some basic augmentation operations to graphs.

Previous work~\cite{xuan2019subgraph} has shown that structural feature expansion can effectively improve the performance of graph classification.
In this paper, 
we take an effective approach to study data augmentation on graphs, as visualized in Fig.~\ref{fig:framework}, and develop four graph augmentation methods, called {\em{random mapping}}, {\em{vertex-similarity mapping}}, {\em{motif-random mapping}} and {\em{motif-similarity mapping}}, respectively. 
The idea is to generate more virtual data for small datasets via heuristic modification and transformation of graph structures. 
Since the generated graphs are artificial and treated as weakly labeled data, their availability remains to be verified.
Therefore, we introduce a concept of label reliability, which reflects the matching degree between examples and their labels against classifier, to filter fine augmented examples from the generated data.
Furthermore, we introduce a model evolution framework, named \textit{M-Evolve}, which combines graph augmentation, data filtration and model retraining to optimize classifiers. We demonstrate that the new framework achieves a significant improvement of performance on graph classification.
The main contributions of this work are summarized as follows:
\begin{itemize}
  \item 
  We utilize the technique of data augmentation on graph classification, and develop several methods to generate effective weakly labeled data for graph benchmark datasets.
  To the best of our knowledge, this is the first work that uses data augmentation in graph mining.
  \item
  We establish a generic model evolution framework named \textit{M-Evolve} for enhancing graph classification, which can be easily combined with existing graph classification models, and we also optimize their performances.
  \item
  We conduct experiments on six benchmark datasets.
  Experimental results demonstrate the superiority of our \textit{M-Evolve} framework in helping five graph classification algorithms to achieve significant improvement of performances.
  
\end{itemize}

The rest of paper is organized as follows. First, in Sec.~\ref{sec:RELATED-WORK}, we review the related works on graph classification. 
Then, in Sec.~\ref{sec:Method}, we introduce and analyze several graph augmentation methods and a new model evolution framework.
Thereafter, we present extensive experiments in Sec.~\ref{sec:experiment}, with detailed discussions.
Finally, we conclude the paper and outline future work in Sec.~\ref{sec:Conclusion}.

\section{Related work} \label{sec:RELATED-WORK}
\subsection{Graph Classification}
\subsubsection{Graph Kernel Methods}
Graph kernels perform graph comparison by recursively decomposing pairwise graphs from the dataset into atomic substructures
and using a similarity function among these substructures. 
Intuitively, graph kernels can be understood as functions measuring the similarity of pairwise graphs.
%
Generally, the kernels can be designed by considering various structural properties like 
the similarity of local neighborhood structures (WL kernel~\cite{shervashidze2011weisfeiler}, propagation kernel~\cite{neumann2016propagation}),
the occurrence of certain graphlets or subgraphs (graphlet kernel~\cite{shervashidze2009efficient}),
the number of walks in common (random walk kernel~\cite{gartner2003graph, kashima2003marginalized, mahe2004extensions, sugiyama2015halting}),
and the attributes and lengths of the shortest paths (SP kernel~\cite{borgwardt2005shortest}).

\subsubsection{Embedding Methods}
Graph embedding methods~\cite{cai2018comprehensive, chen2020graph, fu2019nes, guo2019unsupervised} capture the graph topology and derive a fixed number of features,  ultimately achieving vector representations for the whole graph.
For prediction on the graph level, this approach is compatible with any standard machine learning classifier such as support vector machine (SVM), $k$-nearest neighbors and random forest.
Widely used embedding methods include graph2vec~\cite{narayanan2017graph2vec}, structure2vec~\cite{dai2016discriminative}, subgraph2vec~\cite{narayanan2016subgraph2vec}, etc.

\subsubsection{Deep Learning Methods}
Recently, increasing attention is drawn to the application of deep learning to graph mining 
and a wide variety of graph neural network (GNN) frameworks have been proposed for graph classification, including methods inspired by CNN, RNN, etc.
One typical approach is to obtain a representation of the entire graph by aggregating the vertex embeddings that are the output of GNNs~\cite{duvenaud2015convolutional, gilmer2017neural}. 
Some sequential methods~\cite{li2015gated, jin2018learning, you2018graphrnn} handle these graphs with varying sizes by transforming them into sequences of fixed-length vectors and then feeding them to RNN.
In addition, some hierarchical clustering methods~\cite{defferrard2016convolutional, simonovsky2017dynamic, fey2018splinecnn} learn hierarchical graph representations by combining GNNs with clustering algorithms.
Notably, some recent works design universal graph pooling modules, which can learn the hierarchical representations of graphs and are compatible with various GNN architectures, 
e.g., 
DiffPool~\cite{ying2018hierarchical} learns a differentiable soft cluster assignment for vertices and then maps them to a coarsened graph layer by layer, and EigenPool~\cite{ma2019graph} compresses the vertex features and local structures into coarsened signals via graph Fourier transform.

\begin{table}[!t]
    \renewcommand\arraystretch{1.2}
    \centering
    \caption{Terms and notations used in this paper.}
	\label{tb:notation}
    \resizebox{\linewidth}{!}{%
    \begin{tabular}{l r} 
    \toprule[1pt]
    Symbol               & Definition                                      \\ 
    \midrule[0.5pt]
	$G, G'$                					& Original/augmented graph                   \\
	$V, E$                 					& Sets of vertices/edges in graph $G$              \\
	$m, n$									& Number of edges/vertices \\
	$A$                    					& Adjacency matrix of graph $G$                     \\
	$l$                    					& Length of path                                    \\
	$h_{ij}^l$                 				& Length-$l$ path between vertices $i,j$            \\
	$D, D_{\textit{train}}, D_{\textit{test}}, D_{\textit{val}}$   & Dataset, training/testing/validation set               \\
	$D_{\textit{train}}'$					& Augmented set                                     \\
	$\beta$								    & Budget of edge modification               \\
	$f$                    					& Mapping                                           \\
	$E_{\textit{add}}^c,E_{\textit{del}}^c$ & Candidate edges set of addition/deletion          \\
	$E_{\textit{add}}, E_{\textit{del}}$    & Edges set of addition/deletion                	\\
	$s_{ij}$                  				& Similarity score of $(i,j)$                       \\
	$w_{ij}$                  				& Sampling weight of $(i,j)$                    \\
	$S$                    					& Set of similarity scores                      \\
	$W_{\textit{add}}, W_{\textit{del}}$    & Addition/deletion weights               \\
	$\mathbf{p}$							& Prediction probability vector of example\\
	$\mathbf{q}$  							& Average probability vector of class \\
	$\mathbf{Q}$							& Probability confusion matrix \\
	$r$										& Label reliability          \\
	$\theta$								& Label reliability threshold \\
	$T$                                     & Number of iterations      \\
    \bottomrule[1pt]
    \end{tabular}}
\end{table}

\section{Methodology} \label{sec:Method}
In this section, we first formulate the problem of data augmentation on graphs, and then preset several graph augmentation strategies, which are heuristic and especially fit for graph classification task. The notations used in this paper are listed in TABLE \ref{tb:notation}.

\subsection{Problem Formulation}\label{problem_formulation}
Let $G=(V,E)$ be an undirected and unweighted graph, which consists of a vertex set $V=\left\{v_{i} \mid i=1, \ldots, n\right\}$ and an edge set $E=\left\{e_{i} \mid i=1, \ldots, m\right\}$.
The topological structure of graph $G$ is represented by an $n \times n$ adjacency matrix $A$ with $A_{ij} = 1$ if $(i,j) \in E$ and $A_{ij}=0$ otherwise.
Given pairwise vertices $(i,j)$, the length-$l$ path between them is represented as an ordered edge sequence, i.e., $h_{ij}^{l} = [(v_i, v_1),(v_1, v_2),\ldots, (v_{l-1}, v_j)]$.
Dataset that contains a series of graphs is denoted as $D=\{(G_{i}, y_{i}) \mid i=1, \ldots , t\}$, where $y_i$ is the label of graph $G_i$.
For $D$, an upfront split will be applied to yield disjoint training, validation and testing sets, denoted as $D_{\textit {train}}$, $D_{\textit {val}}$ and $D_{\textit {test}}$, respectively.
The original classifier $C$ will be pre-trained using $D_{\textit {train}}$ and $D_{\textit {val}}$.

We further explore the data augmentation technique for graph classification from a heuristic approach and consider optimizing graph classifiers.
Fig.~\ref{fig:framework} demonstrates the application of data augmentation to graph structured data, which consists of two phases: graph augmentation and data filtration. 
Specifically, we aim to update a classifier with augmented data, which are first generated via graph augmentation and then filtered in terms of their label reliability. 
During graph augmentation, our purpose is to map the graph $G \in D_{\textit{train}}$ to a new graph $G'$ with the formal format, $f:(G,y) \mapsto (G',y)$, where $y$ is the label of $G$. 
We treat the generated graphs as weakly labeled data and classify them into two groups via a label reliability threshold learnt from $D_{\textit{val}}$.
Then, the augmented set $D_{\textit{train}}'$ filtered from the generated graph pool $D_{\textit{pool}}$ will be merged with $D_{\textit {train}}$ to produce the training set:
\begin{equation}
  D_{\textit{train}}^{\textit{new}} = D_{\textit{train}} + D_{\textit{train}}'\ , \quad
  D_{\textit{train}}' \subset D_{\textit{pool}} \ .
\end{equation}
Finally, we finetune or retrain the classifier with $D_{\textit{train}}^{\textit{new}}$, and evaluate it on the testing set $D_{\textit {test}}$.

\subsection{Graph Augmentation}
Graph augmentation aims to expand training data via artificially creating more reasonable virtual data from a limited set of graphs. 
In this paper, we consider augmentation as a topological mapping, which is conducted via heuristic modification or transformation of graph structures.
In order to ensure the approximate reasonability of the generated virtual data, our graph augmentation will follow the following principles:
1) edge modification, where $G'$ is a partially modified graph with some of the edges added/removed from $G$;
2) structure property preservation, where augmentation operation keeps the graph connectivity and the number of edges constant.
Notably, keeping the number of edges is just an optional constraint condition of the graph augmentation, which is convenient for us to adjust parameters.

During edge modification, those edges removed from graph are sampled from the candidate edge set $E_{\textit{del}}^{\textit{c}}$, while the edges added to the graph are sampled from the candidate pairwise vertices set $E_{\textit{add}}^{\textit{c}}$. 
The construction of candidate sets varies for different methods, as further discussed below.

\subsubsection{Random Mapping}
Here, consider {\em{random mapping}} as a simple baseline method. For a given graph $G$, one can randomly remove some edges from $E_{\textit{del}}^{\textit{c}}$ and then link the same number of pairwise vertices, which exist in  $E_{\textit{add}}^{\textit{c}}$.
In this random scenario, the candidate sets are denoted as follows:
\begin{equation}\label{eq:random-candidate}
	\begin{aligned}
		&E_{\textit{del}}^{\textit{c}} =  E , \\
		&E_{\textit{add}}^{\textit{c}} = \{(v_i, v_j) \mid A_{ij}=0;  \ i \neq j \} \subset (V \times V).
	\end{aligned}
\end{equation}

\begin{algorithm}[htb]
    \caption{Vertex-Similarity mapping}  
    \LinesNumbered  
    \label{alg:vs-mapping} 
	\KwIn{Target network $G$, proportion of modification $\beta$.}  
    \KwOut{Augmented graph $G'$}  		
	Get $E_{\textit{add}}^{\textit{c}}$ and $E_{\textit{del}}^{\textit{c}}$  via Eq.~(\ref{eq:random-candidate}) \;
	Compute the addition weights $W_{\textit{add}}$ via Eq.~\ref{eq:cal-add-weight} \;
	$E_{\textit{add}} \leftarrow \textsf{weightRandomSample}(E_{\textit{add}}^{\textit{c}}, \lceil m \cdot \beta \rceil, W_{\textit{add}})$ \;
	Compute the deletion weights $W_{\textit{del}}$ via Eq.~\ref{eq:cal-delete-weight} \;
	$E_{\textit{del}} \leftarrow \textsf{weightRandomSample}(E_{\textit{del}}^{\textit{c}}, \lceil m \cdot \beta \rceil, W_{\textit{del}})$ \;
    Get augmented graph $G'$ via Eq.~(\ref{eq:graph-update}) \;
    \textbf{end} \;
    \textbf{return} $G'$;  
\end{algorithm} 

Notably, in Eq.~(\ref{eq:random-candidate}), $E_{\textit{del}}^{\textit{c}}$ is actually the edge set of the graph, and $E_{\textit{add}}^{\textit{c}}$ is the set of virtual edges which consists of unlinked pairwise vertices.
Then, one can get the set of edges added/removed from $G$ via sampling in the candidate sets randomly:
\begin{equation}\label{eq:random-edgeset}
	\begin{aligned}
		&E_{\textit{del}} = \{e_i \mid i=1, \ldots, \lceil m \cdot \beta \rceil \} \subset E_{\textit{del}}^{\textit{c}} \ , \\
		&E_{\textit{add}} = \{e_i \mid i=1, \ldots, \lceil m \cdot \beta \rceil \} \subset E_{\textit{add}}^{\textit{c}} \ ,
	\end{aligned}
\end{equation}
where $\beta$ is the budget of edge modification and $\lceil x \rceil = \textbf{ceil}(x)$.
Finally, based on the {\em{random mapping}}, the connectivity structure of the original graph is modified to generate a virtual graph:
\begin{equation}\label{eq:graph-update}
	G' = (V, (E \cup E_{\textit{add}}) \backslash E_{\textit{del}}) \ .
\end{equation}

\subsubsection{Vertex-Similarity Mapping}
Vertex similarity, which reflects the number of common features shared by a pair of vertices, has been widely applied to graph mining tasks such as link prediction and community detection.
Empirically, vertices in graph tend to intersect with each other due to their high similarity. 
Therefore, this similarity index is used to link those vertices of high similarity for graph augmentation.

\begin{figure}[!t]
	\centering
	\includegraphics[width=\linewidth]{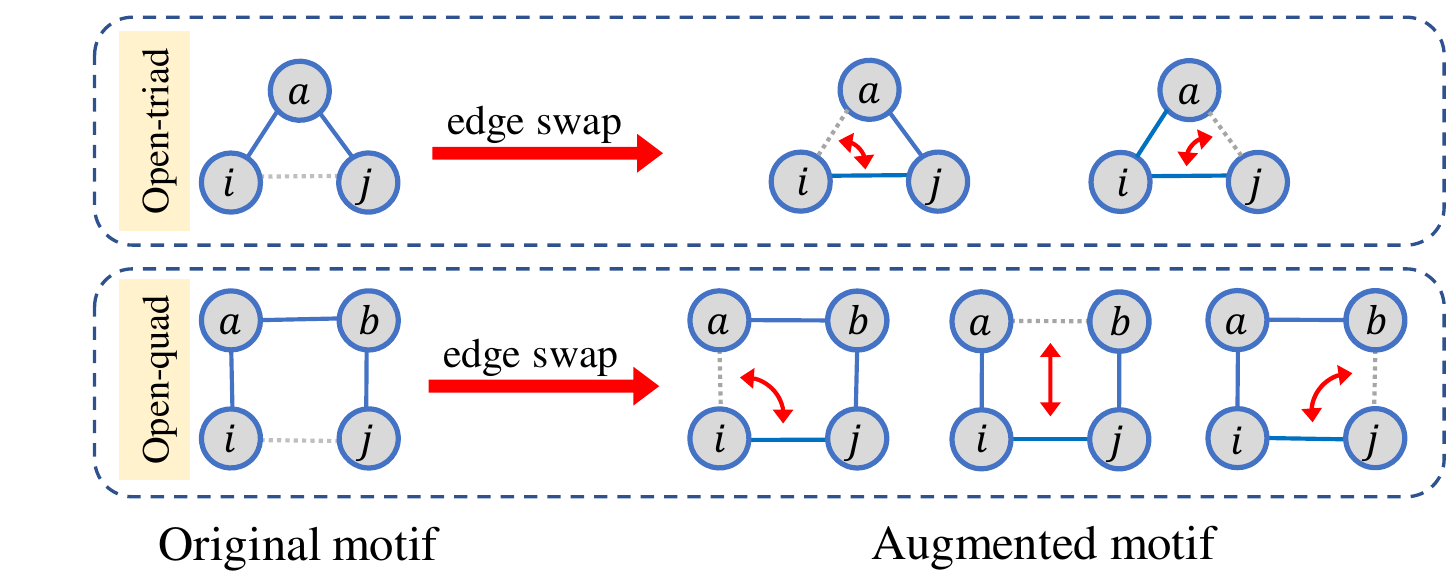}
	\caption{An example for edge swapping in motif-random mapping. {\em{Left}} : Common graph motifs such as open-triad and open-quad. {\em{Right}} : Augmented motifs obtained via edge swapping. The dashed lines represent the virtual edges in graph.}
	\label{fig:motif-edge-swapping}
\end{figure}

\begin{figure*}[!t]
\centering
\includegraphics[width=\textwidth]{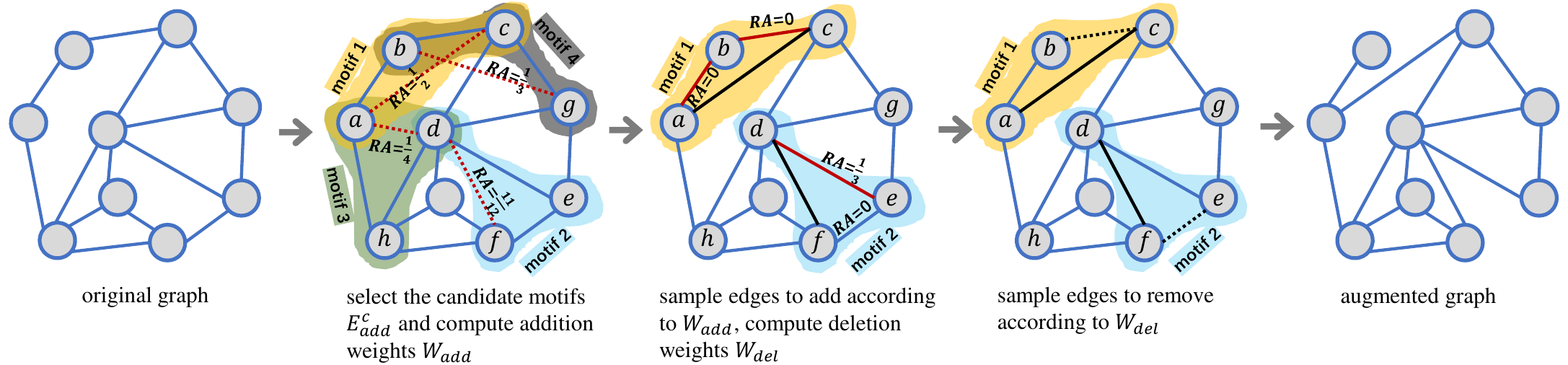}
\caption{An example for motif-similarity mapping. Red lines, both dashed and solid, represent the candidates. Black lines, both dashed and solid, represent the modified edges.}
\label{fig:similarity-mapping}
\end{figure*}

{\em{Vertex-similarity mapping}} shares the same candidate sets with {\em{random mapping}} but conducts edge selection with weighted random sampling.
In other words, {\em{random mapping}} makes selections with equal probability, while {\em{vertex-similarity mapping}} assigns all entries in $E_{\textit{add}}^{\textit{c}}$ and $E_{\textit{del}}^{\textit{c}}$ with relative weights that are associated with the vertex similarity scores. 
Specifically, before sampling, the similarity scores are computed over all entries in $E_{\textit{add}}^{\textit{c}}$.
Since edge modification focuses on optimizing local structures, we compute the vertex similarity scores using the {\emph{Resource Allocation (RA)}} index~\cite{zhou2009predicting} which has been proven its superiority among several local similarity measures in~\cite{zhou2009predicting}.
For each entry $(v_i, v_j)$ in $E_{\textit{add}}^{\textit{c}}$, the {\emph{RA}} score $s_{ij}$ and addition weight $w_{i j}^{\textit{add}}$ are computed as follows:
\begin{equation}\label{eq:cal-add-weight}
	\begin{aligned}
		&s_{i j}=\sum_{z \in \Gamma(i) \cap \Gamma(j)} \frac{1}{d_z}, \ S=\{s_{i j} \mid \forall(v_i, v_j) \in E_{\textit{add}}^{\textit{c}} \},\\
		&w_{i j}^{\textit{add}}=\frac{s_{i j}}{\sum_{s \in S} s}, \ W_{\textit{add}}=\{w_{i j}^{\textit{add}} \mid \forall(v_i, v_j) \in E_{\textit{add}}^{\textit{c}} \} ,
	\end{aligned}
\end{equation}
where $\Gamma(i)$ denotes the one-hop neighbors of $i$ and $d_z$ denotes the degree of vertex $z$. 
Weighted random sampling means that the probability for an entry in $E_{\textit{add}}^{\textit{c}}$ to be selected is proportional to its addition weight $w_{i j}^{\textit{add}}$.
It is worth noting that many other similarity indices can also be applied into this scheme.
Similarly, during edge deletion, the probability of edge sampled from $E_{\textit{del}}^{\textit{c}}$ is proportional to the deletion weight $w_{ij}^{\textit{del}}$ as follows:
\begin{equation}\label{eq:cal-delete-weight}
	w_{ij}^{\textit{del}} = 1-\frac{s_{i j}}{\sum_{s \in S} s}, \ W_{\textit{del}}=\{w_{i j}^{\textit{del}} \mid \forall(v_i, v_j) \in E_{\textit{del}}^{\textit{c}} \} ,
\end{equation}
which means that these edges with smaller {\emph{RA}} scores have more chances to be removed.

\subsubsection{Motif-Random Mapping}
Graph motifs are sub-graphs that repeat themselves in a specific graph or even among various graphs. 
Each of these sub-graphs, defined by a particular pattern of interactions between vertices, may describe a framework in which particular functions are accomplished efficiently.
In this paper, only those open motifs with a chain structure are considered, as shown in Fig.~\ref{fig:motif-edge-swapping}, so that the motif discovery process can be replaced by path search. 
For instance, common motifs such as open-triad is equivalent to length-2 paths emanating from the head vertex that induce a triangle, which gives the following equivalence relation:
\begin{equation}
	\begin{aligned}
		&\textbf{open-triad} \equiv [(v_i,v_a),(v_a,v_j)] \Rightarrow l = 2 , \\
		&\textbf{open-quad} \equiv [(v_i,v_a),(v_a,v_b),(v_b,v_j)] \Rightarrow l = 3 ,
	\end{aligned}
\end{equation}
where $l$ is the length of motif.

The {\em{motif-random mapping}} aims to finetune these motifs to the approximately equivalent ones via edge swapping.
As shown in Fig.~\ref{fig:motif-edge-swapping}, during edge swapping, edge addition operation takes effect between the head and the tail vertices of the motif, while edge deletion operation randomly removes an edge in the motif.
For all length-$l$ motifs, the candidate set of edge addition is:
\begin{equation}\label{eq:motif-mapping-cand-add}
	E_{\textit{add}}^{\textit{c}} = \{(v_i, v_j) \mid A_{ij}=0, A_{ij}^{l} \neq 0 \},
\end{equation}
\begin{algorithm}[htb]
    \caption{Motif-Similarity mapping}  
    \LinesNumbered  
    \label{alg:ms-mapping} 
	\KwIn{Target network $G$, length of motif $l$, proportion of modification $\beta$.}  
    \KwOut{Augmented graph $G'$}  		
	Get $E_{\textit{add}}^{\textit{c}}$ via Eq.~(\ref{eq:motif-mapping-cand-add}) \;
	Compute the addition weights $W_{\textit{add}}$ via Eq.~\ref{eq:cal-add-weight} \;
	$E_{\textit{add}} \leftarrow \textsf{weightRandomSample}(E_{\textit{add}}^{\textit{c}}, \lceil m \cdot \beta \rceil, W_{\textit{add}})$ \;
	Initialize $E_{\textit{del}} = \emptyset$  \;
	\For{${\textit{each}} (v_i,v_j) \in E_{\textit{add}} $}
	{
		Get the length-$l$ motif $h_{ij}^{l}$ via path search: $h_{ij}^{l} \leftarrow \textsf{pathSearch}(i,j,l)$ \;
		Compute the deletion weights $W_{\textit{del}}$ via Eq.~\ref{eq:cal-delete-ms} \;
		$e_{\textit{del}} \leftarrow \textsf{weightRandomSample}(h_{ij}^{l}, 1, W_{\textit{del}})$ \;	
		Add $e_{\textit{del}}$ to $ E_{\textit{del}}$ \;
	}
    Get augmented graph $G'$ via Eq.~(\ref{eq:graph-update}) \;
    \textbf{end} \;
    \textbf{return} $G'$;  
\end{algorithm}
where $A^{l}$ is $A$ to the power of $l$. Then, one can get $E_{\textit{add}}$, the set of edges added to $G$, via sampling from $E_{\textit{add}}^{\textit{c}}$ randomly.
For each pair of vertices $(v_i, v_j)$ in $E_{\textit{add}}$, there exists a length-$l$ motif $h_{ij}^{l}$, which has head vertex $v_i$ and tail vertex $v_j$.
Next, we randomly sample one edge in $h_{ij}^{l}$ to remove and all of these removed edges constitute $E_{\textit{del}}$. 
Finally, the virtual graph can be obtained via Eq.~(\ref{eq:graph-update}).

\subsubsection{Motif-Similarity Mapping}
{\em{Motif-similarity mapping }}inherits from {\em{motif-random mapping}} and conducts edge sampling in term of vertex similarity.
Specifically, for the candidate set of edge addition given in Eq.~(\ref{eq:motif-mapping-cand-add}), the probability for an arbitrary entry to be selected is proportional to its addition weight $w_{ij}^{\textit{add}}$ according to Eq.~(\ref{eq:cal-add-weight}).
Similarly, during edge deletion, the probability of edge sampled from $h_{ij}^{l}$ is proportional to the deletion weight $w_{ij}^{\textit{del}}$ as follows:
\begin{equation}\label{eq:cal-delete-ms}
	w_{ij}^{\textit{del}} = 1-\frac{s_{i j}}{\sum_{s \in S} s}, \ W_{\textit{del}}=\{w_{i j}^{\textit{del}} \mid \forall(v_i, v_j) \in h_{ij}^{l} \} .
\end{equation}
The whole process for {\em{motif-similarity mapping}} with $l=2$ and $\lceil m \cdot \beta \rceil = 2$ is illustrated in Fig.~\ref{fig:similarity-mapping}.

\subsection{Data Filtration}
\begin{figure}[htb]
	\centering
	\includegraphics[width=\linewidth]{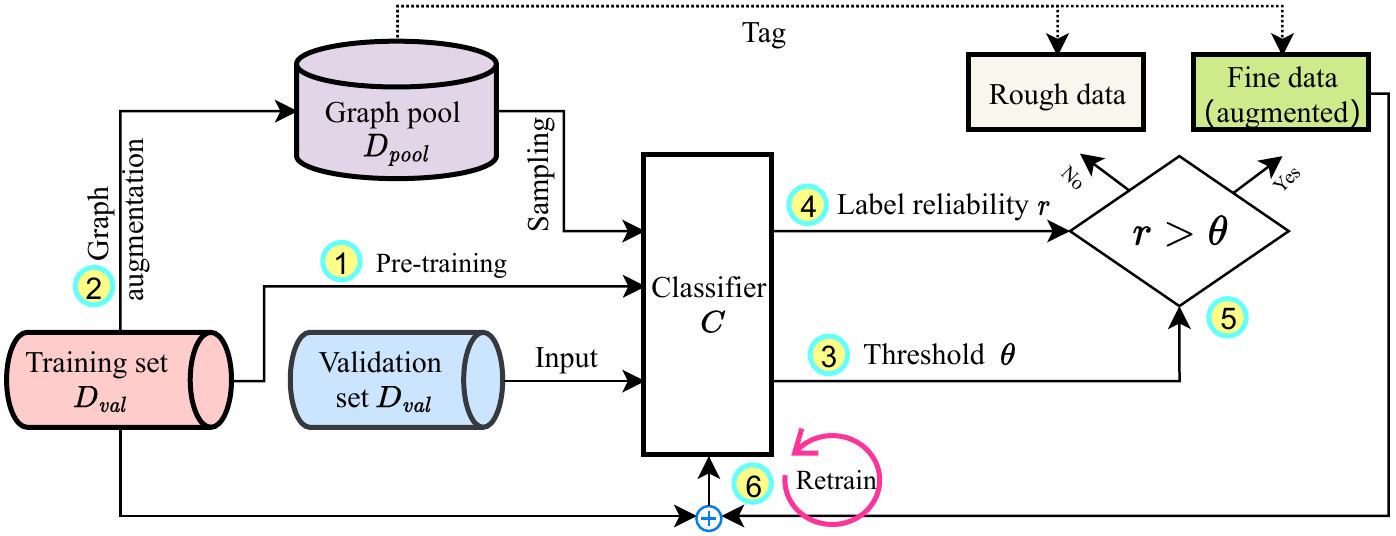}
	\caption{The architecture of the model evolution. The complete workflow proceeds as follows: 
			 1) pre-train graph classifier using training set; 
			 2) apply graph augmentation to generate data pool; 
			 3) compute the label reliability threshold using validation set; 
			 4) compute the label reliability of examples sampled from graph pool;
			 5) filter data and obtain augmented set using threshold;
			 6) retrain graph classifier using the union of training set and augmented set.}
	\label{fig:model-evolution-farmework}
\end{figure}
In computer vision, the technique of data augmentation is widely used to generate augmented examples, which can be directly treated as new training data.
For instance, after an image of cat undergoes simple data augmentation such as geometric transformation or color depth adjustment, the resulting new image retains the explicit semantics and is still an image of cat in human vision.
However, due to the difference between image data and graph structured data, the examples generated via graph augmentation may lose original semantics.
By assigning the label of the original graph to the generated graph directly during graph augmentation, one cannot determine whether the assigned label is reliable.
Therefore, the concept of label reliability is employed here to measure the matching degree between examples and labels against classifier.

Each graph $G_i$ in $D_{\textit{val}}$ will be fed into classifier $C$ to obtain the prediction vector $\mathbf{p}_i \in \mathbb{R}^{|Y|}$, which represents the probability distribution of how likely an input example belongs to each possible class, and $|Y|$ is the number of classes for labels. 
Then, a probability confusion matrix $\mathbf{Q} \in \mathbb{R}^{|Y|\times |Y|}$, in which the entry $q_{ij}$ represents the average probability that the classifier classifies the graphs of the $i$-th class into the $j$-th class, is computed as follows:
\begin{equation}\label{eq:confusion-matrix}
	\begin{aligned}
		&\mathbf{q}_k = [q_{k1}, q_{k2}, \ldots, q_{k|Y|}]^\top = \frac{\sum_{y_i=k} \mathbf{p}_i}{\Omega_k} \ , \\
		&\mathbf{Q} = [\mathbf{q}_1, \mathbf{q}_2, \ldots, \mathbf{q}_{|Y|}] \ ,
	\end{aligned}
\end{equation}
where $\Omega_k$ is the number of graphs belonging to the $k$-th class in $D_{\textit{val}}$ and $\mathbf{q}_k$ is the average probability distribution of the $k$-th class.

The label reliability of an example $(G_i, y_i)$ is defined as the product of example probability distribution $\mathbf{p}_i$ and class probability distribution $\mathbf{q}_{y_i}$ as follows:
\begin{equation}\label{eq:label-reliability}
  r_i = {\mathbf{p}_i}^\top \mathbf{q}_{y_i} .
\end{equation}
Notably, the definition indicates that the example which can be predicted correctly by classifier with a greater probability tends to have higher label reliability.

A threshold $\theta$ used to filter the generated data is defined as:
\begin{equation}\label{eq:threshold}
  \theta = \arg \min_\theta \sum_{(G_i, y_i) \in D_{\textit{val}}} \Phi[(\theta-r_i) \cdot g(G_i, y_i)] \ ,
\end{equation} 
where $g(G_i, y_i)=1$ if $C(G_i)=y_i$ and $g(G_i, y_i)=-1$ otherwise, and $\Phi(x)=1$ if $x>0$ and $\Phi(x)=0$ otherwise.
\begin{algorithm}[htb]
	\caption{{\em{M-Evolve}}}  
	\LinesNumbered  
	\label{alg:model-evolution} 
	\KwIn{Training set $D_{\textit{train}}$, validation set $D_{\textit{val}}$, graph augmentation $f$, number of iterations $T$.}  
	\KwOut{Evolutive model $C'$}  	
	Pre-training classifier $C$ using $D_{\textit{train}}$ and $D_{\textit{val}}$ \;
	Initalize $\textit{iteration} = 0$\;
	\For{$\textit{iteration} < T$}
	{
	  Graph augmentation: $D_\textit{pool} \leftarrow f(D_{\textit{train}})$ \;
	  For all graphs $G_i$ in $D_{\textit{val}}$ classified by $C$, get $\mathbf{p}_i$ \;
	  Get probability confusion matrix $\mathbf{Q}$ via Eq.~\ref{eq:confusion-matrix} \;
	  For all graphs $G_i$ in $D_{\textit{val}}$ classified by $C$, get $r_i$ via Eq.~\ref{eq:label-reliability}\;
	  Get the label reliability threshold $\theta$ via Eq.~\ref{eq:threshold} \;
	  For all samples $(G_i, y_i)$ in $D_{\textit{pool}}$ classified by $C$, compute $r_i$, \textbf{if} $r_i > \theta$, $D_{\textit{train}}.append((G_i, y_i))$ \;
	  Get evolutive classifier: $C' \leftarrow \mathsf{retrain}(C ,D_{\textit{train}})$ \;
	  $\textit{iteration} \leftarrow \textit{iteration} + 1$ \;
	  $C \leftarrow C'$ \;
	}
	\textbf{end} \;
	\textbf{return} $C'$;  
\end{algorithm} 
\vspace{-12pt}

\begin{table}[htp]
	\renewcommand\arraystretch{1}
	\centering
	\huge
	\caption{Dataset properties. $|D|$ is the number of graphs in dataset $D$, $|Y|$ is the number of classes for labels, $\mathit{Avg.}|V|$ is the average number of vertices, $\mathit{Avg.}|E|$ is the average number of edges, $\mathit{Min./Max.}$ is the minimal/maximal scale of graphs in dataset and $\mathit{bias}$ is the proportion of the dominant class.}
	\label{tb:dataset}
	\resizebox{\linewidth}{!}{%
	\begin{tabular}{ccccllc} 
	\toprule
	Collections                                                                     & Dataset  & \multicolumn{1}{l}{$|D|$ } & \multicolumn{1}{l}{$|Y|$ } & \multicolumn{1}{l}{$\textit{Avg.}|V|\sim(\textit{Min./Max.})$ } & \multicolumn{1}{l}{$\textit{Avg.}|E|\sim(\textit{Min./Max.})$ }  & \multicolumn{1}{l}{\textit{bias} (\%) }  \\ 
	\midrule
	\multirow{3}{*}{\begin{tabular}[c]{@{}c@{}}Chemical ~\\Compounds \end{tabular}} & MUTAG    & 188                        & 2                          & 17.93~$\sim$~(10/28)                            & 19.79~$\sim$~(10/33)                             & 66.5                                    \\
																					& PTC-MR   & 344                        & 2                          & 14.29~$\sim$~(3/64)                             & 14.69~$\sim$~(2/71)                              & 55.8                                    \\
																					& ENZYMES  & 600                        & 6                          & 32.63~$\sim$~(3/125)                            & 62.14~$\sim$~(3/149)                             & 16.7                                    \\ 
	\midrule
	\multirow{3}{*}{Brain}                                                          & KKI      & 83                         & 2                          & 26.96~$\sim$~(5/90)                             & 48.42~$\sim$~(4/237)                             & 55.4                                    \\
																					& Peking-1 & 85                         & 2                          & 39.31~$\sim$~(7/134)                            & 77.35~$\sim$~(7/535)                             & 57.6                                    \\
																					& OHSU     & 79                         & 2                          & 82.01~$\sim$~(9/171)                            & 199.66~$\sim$~(9/823)                            & 55.7                                    \\
	\bottomrule
	\end{tabular}}
\end{table}
The process of threshold optimization tends to ensure that correctly classified examples have greater label reliability than incorrectly classified examples.
\subsection{Model Evolution Framework}
Model evolution aims to optimize classifiers via graph augmentation, data filtration and model retraining iteratively, and ultimately improve the performance on graph classification.
Fig.~\ref{fig:model-evolution-farmework} and Algorithm~\ref{alg:model-evolution} demonstrate the workflow and the procedure of {\em{M-Evolve}}, respectively.
Here, a variable, number of iterations $T$, is introduced for repeating the above workflow to continuously augment the dataset and optimize classifier.

\section{Experiments} \label{sec:experiment}
\subsection{Datasets} \label{Datasets}
We evaluate the proposed methods against six benchmark datasets from bioinformatics and chemoinformatics: Mutag~\cite{debnath1991structure}, PTC-MR~\cite{helma2001predictive}, ENZYMES~\cite{borgwardt2005protein}, KKI, Peking-1 and OHSU~\cite{pan2016task}.
The first three represent the graph collections of chemical compounds in which vertices correspond to molecular structures and edges indicate chemical bonds between them.
And the last three are constructed from brain networks, where vertices correspond to the Regions of Interest (ROI) and edges represent the correlations between two ROIs.
The specifications of datasets are given in TABLE~\ref{tb:dataset}.

\begin{table*}[ht]
	\renewcommand\arraystretch{1.2}
	\centering
	\caption{Graph classification results of original and evolutive models. The best results are marked in bold.
			 The far-right column gives the average relative improvement rate (Avg RIMP) in accuracy.}
	\label{tb:result}
	\resizebox{\textwidth}{!}{%
	\begin{tabular}{c|c|cccc|cccc|cccc|cccc|c|c} 
	\hline
	\multirow{3}{*}{Dataset}  & \multirow{3}{*}{Mapping} & \multicolumn{17}{c|}{Graph Classification Model}                                                                                                                                                                                                                                            & \multirow{3}{*}{Avg
	RIMP}  \\ 
	\cline{3-19}
							  &                          & \multicolumn{4}{c|}{SF}                 & \multicolumn{4}{c|}{NetLSD}             & \multicolumn{4}{c|}{Graph2vec}          & \multicolumn{4}{c|}{Gl2vec}             & \multirow{2}{*}{Diffpool} &                           \\ 
	\cline{3-18}
							  &                          & SVM             & Log             & KNN             & RF              & SVM             & Log             & KNN             & RF              & SVM             & Log             & KNN             & RF              & SVM             & Log             & KNN             & RF              &                           &                           \\ 
	\hline
	\multirow{5}{*}{MUTAG}    & original                 & 0.822           & 0.824           & 0.824           & 0.846           & 0.823           & 0.829           & 0.828           & 0.836           & 0.737           & 0.820           & 0.784           & 0.820           & 0.746           & 0.830           & 0.800           & 0.817           & 0.801                     & --                        \\
							  & random                   & 0.843           & 0.845           & 0.846           & 0.878           & 0.855           & 0.851           & 0.860           & 0.886           & 0.756           & 0.844           & 0.793           & 0.847           & 0.748           & 0.851           & 0.820           & 0.841           & 0.810                     & 2.78\%                    \\
							  & vertex-similarity        & 0.848           & 0.855           & 0.840           & 0.870           & 0.843           & 0.845           & 0.856           & 0.874           & 0.750           & 0.850           & 0.801           & \textbf{0.861}  & 0.748           & 0.845           & 0.823           & \textbf{0.850}  & 0.825                     & 2.86\%                    \\
							  & motif-random             & 0.860           & 0.852           & 0.846           & 0.887           & 0.861           & 0.862           & 0.856           & 0.882           & \textbf{0.761}  & 0.850           & 0.803           & 0.859           & 0.752           & 0.856           & 0.829           & 0.845           & 0.807                     & 3.63\%                    \\
							  & motif-similarity         & \textbf{0.863} & \textbf{0.855}  & \textbf{0.850}  & \textbf{0.890}   & \textbf{0.861}  & \textbf{0.864}  & \textbf{0.861}  & \textbf{0.892}  & 0.759           & \textbf{0.851}  & \textbf{0.809}  & 0.852           & \textbf{0.762}  & \textbf{0.863}  & \textbf{0.833}  & 0.846           & \textbf{0.831}            & \textbf{4.00\%}           \\ 
							  \hline
	\multirow{5}{*}{PTC-MR}   & original                 & 0.551           & 0.566           & 0.577           & 0.587           & 0.543           & 0.578           & 0.548           & 0.576           & 0.571           & 0.518           & 0.509           & 0.549           & 0.572           & 0.538           & 0.507           & 0.550           & 0.609                     & --                        \\
							  & random                   & 0.611           & 0.590           & 0.605           & 0.618           & 0.579           & 0.580           & 0.590           & 0.607           & 0.580           & 0.572           & 0.547           & 0.592           & 0.587           & 0.571           & 0.528           & 0.594           & 0.637                     & 5.77\%                    \\
							  & vertex-similarity        & 0.595           & 0.594           & 0.601           & 0.622           & 0.577           & 0.580           & 0.578           & 0.601           & 0.592           & 0.571           & 0.548           & \textbf{0.599}  & \textbf{0.601}  & 0.575           & \textbf{0.554}  & \textbf{0.605}  & 0.636                     & 6.22\%                    \\
							  & motif-random             & 0.615           & 0.595           & 0.609           & \textbf{0.630}  & \textbf{0.595}  & 0.583           & 0.582           & 0.612           & 0.587           & 0.570           & 0.551           & 0.592           & 0.587           & 0.578           & 0.535           & 0.596           & 0.627                     & 6.37\%                    \\
							  & motif-similarity         & \textbf{0.616}  & \textbf{0.595}  & \textbf{0.610}  & 0.624           & 0.581           & \textbf{0.583}  & \textbf{0.597}  & \textbf{0.620}  & \textbf{0.596}  & \textbf{0.579}  & \textbf{0.553}  & 0.593           & 0.588           & \textbf{0.579}  & 0.545           & 0.602           & \textbf{0.639}            & \textbf{6.97\%}           \\ 
							  \hline
	\multirow{5}{*}{ENZYMES}  & original                 & 0.309           & 0.237           & 0.287           & 0.397           & 0.337           & 0.287           & 0.304           & 0.349           & 0.361           & 0.253           & 0.283           & 0.337           & \textbf{0.348}  & 0.268           & 0.238           & 0.318           & 0.487                     & --                        \\
							  & random                   & 0.347           & 0.412           & 0.302           & 0.412           & 0.353           & 0.287           & 0.327           & 0.369           & 0.336           & 0.269           & 0.290           & 0.346           & 0.286           & 0.273           & 0.259           & 0.350           & 0.500                     & 7.25\%                    \\
							  & vertex-similarity        & \textbf{0.368}  & 0.416           & \textbf{0.333}  & \textbf{0.431}  & 0.352           & \textbf{0.295}  & 0.329           & 0.359           & 0.372 		   & \textbf{0.279}  & 0.290           & \textbf{0.369}  & 0.345           & 0.266           & \textbf{0.274}  & 0.357           & 0.489                     & \textbf{11.12\%}          \\
							  & motif-random             & 0.364           & \textbf{0.418}  & 0.317           & 0.418           & 0.365           & 0.294           & \textbf{0.340}  & 0.367           & \textbf{0.376}  & 0.268           & \textbf{0.299}  & 0.355           & 0.298           & 0.273           & 0.264           & 0.356           & \textbf{0.508}            & 10.20\%                    \\
							  & motif-similarity         & 0.363           & 0.414           & 0.317           & 0.415           & \textbf{0.375}  & 0.291           & 0.335           & \textbf{0.376}  & 0.352           & 0.270           & 0.289           & 0.352           & 0.291           & \textbf{0.280}  & 0.260           & \textbf{0.358}  & 0.506                     & 9.55\%                    \\ 
							  \hline
	\multirow{5}{*}{KKI}      & original                 & 0.550           & 0.500           & 0.520           & 0.517           & 0.548           & 0.524           & 0.512           & 0.496           & 0.549           & 0.527           & 0.524           & 0.552           & 0.538           & 0.502           & 0.526           & 0.502           & 0.523                     & --                        \\
							  & random                   & 0.600           & 0.544           & 0.554           & 0.622           & 0.599           & 0.535           & 0.553           & 0.562           & 0.580           & 0.568           & 0.594           & 0.574           & 0.556           & 0.508           & 0.544           & 0.544           & 0.580                     & 7.97\%                    \\
							  & vertex-similarity        & 0.601           & 0.556           & 0.559           & \textbf{0.654}  & 0.618           & \textbf{0.568}  & \textbf{0.568}  & \textbf{0.585}  & \textbf{0.603}  & 0.570           & 0.573           & 0.655           & 0.578           & \textbf{0.597}  & 0.582           & 0.588           & 0.597                     & 12.87\%                   \\
							  & motif-random             & 0.598           & 0.560           & \textbf{0.578}  & 0.647           & 0.603           & 0.563           & 0.558           & 0.574           & 0.586           & 0.606           & 0.592           & \textbf{0.661}  & 0.567           & 0.593           & 0.596           & \textbf{0.604}  & 0.586                     & 13.11\%                   \\
							  & motif-similarity         & \textbf{0.607}  & \textbf{0.560}  & 0.561           & 0.649           & \textbf{0.619}  & 0.558           & 0.565           & 0.582           & 0.587           & \textbf{0.606}  & \textbf{0.603}  & 0.634           & \textbf{0.581}  & 0.592           & \textbf{0.597}  & 0.582           & \textbf{0.612}            & \textbf{13.36\%}          \\ 
							  \hline
	\multirow{5}{*}{Peking-1} & original                 & 0.578           & 0.548           & 0.541           & 0.558           & 0.605           & 0.612           & 0.589           & 0.591           & 0.572           & 0.522           & 0.474           & 0.522           & 0.555           & 0.522           & 0.521           & 0.521           & 0.586                     & --                        \\
							  & random                   & 0.660           & 0.562           & 0.603           & 0.627           & 0.652           & 0.631           & 0.662           & 0.654           & 0.579           & 0.547           & 0.546           & 0.597           & 0.584           & 0.555           & 0.559           & 0.607           & 0.650                     & 9.20\%                    \\
							  & vertex-similarity        & \textbf{0.672}  & 0.571           & 0.615           & 0.623           & 0.652           & 0.644           & 0.622           & 0.666           & \textbf{0.638}  & 0.557           & 0.562           & 0.627           & \textbf{0.619}  & 0.561           & 0.566           & 0.632           & \textbf{0.657}            & 11.47\%                   \\
							  & motif-random             & 0.681           & 0.583           & 0.619           & 0.644           & 0.668           & 0.636           & 0.666           & 0.689           & 0.579           & 0.553           & \textbf{0.593}  & 0.612           & 0.605           & \textbf{0.581}  & 0.567           & 0.633           & 0.654                     & 12.34\%                   \\
							  & motif-similarity         & 0.670           & \textbf{0.587}  & \textbf{0.624}  & \textbf{0.663}  & \textbf{0.694}  & \textbf{0.648}  & \textbf{0.671}  & \textbf{0.699}  & 0.581           & \textbf{0.565}  & 0.564           & \textbf{0.630}  & 0.607           & 0.563           & \textbf{0.572}  & \textbf{0.635}  & 0.632                     & \textbf{12.72\%}          \\ 
							  \hline
	\multirow{5}{*}{OHSU}     & original                 & 0.610           & 0.595           & 0.610           & 0.667           & 0.547           & 0.489           & 0.549           & 0.581           & 0.557           & 0.577           & 0.585           & 0.567           & 0.557           & 0.541           & 0.544           & 0.557           & 0.543                     & --                        \\
							  & random                   & 0.643           & 0.635           & 0.645           & 0.697           & 0.595           & 0.534           & 0.582           & 0.641           & 0.557           & 0.640           & 0.628           & 0.645           & 0.564           & 0.595           & 0.570           & 0.642           & 0.637                     & 8.08\%                    \\
							  & vertex-similarity        & 0.653           & 0.641           & 0.643           & 0.722           & \textbf{0.641}  & \textbf{0.546}  & \textbf{0.614}  & \textbf{0.661}  & 0.557           & 0.658           & 0.623           & 0.673           & 0.557           & \textbf{0.625}  & 0.625           & 0.632           & \textbf{0.640}            & \textbf{10.82\%}          \\
							  & motif-random             & 0.650           & 0.638           & 0.648           & \textbf{0.728}  & 0.613           & 0.546           & 0.584           & 0.641           & 0.557           & 0.653           & 0.633           & \textbf{0.686}  & 0.564           & 0.602           & 0.625           & 0.645           & 0.627                     & 10.04\%                   \\
							  & motif-similarity         & \textbf{0.656}  & \textbf{0.641}  & \textbf{0.650}  & 0.726           & 0.638           & 0.541           & 0.587           & 0.641           & \textbf{0.557}  & \textbf{0.678}  & \textbf{0.635}  & 0.650           & \textbf{0.572}  & 0.605           & \textbf{0.625}  & \textbf{0.652}  & 0.615                     & 10.33\%         \\
							  \hline
\end{tabular}}
\end{table*}

\subsection{Graph Classification Methods} \label{GC-methods}
We consider the following five graph classification methods in our experiments, the first two are graph embedding, the middle two are kernel models, and the last one is the GNN model.
The implementations of the first four methods are integrated in the package named \texttt{karateclub}\footnote{https://github.com/benedekrozemberczki/karateclub} and are easy to call. The implementation of \texttt{Diffpool} is available online\footnote{https://github.com/RexYing/diffpool}.
\begin{itemize}
\item
\textbf{SF}~\cite{de2018simple}. 
It is an embedding method and performs graph classification by spectral decomposition of the graph Laplacian, i.e., it relies on spectral features of the graph.
\item 
\textbf{Graph2vec}~\cite{narayanan2017graph2vec}.
It extends the document embedding methods to graph classification and learns a distributed representation of the entire graph via document embedding neural networks.
\item 
\textbf{NetLSD}~\cite{tsitsulin2018netlsd}.
It is a kernel method and performs graph classification by extracting compact graph signatures that inherit the formal properties of the Laplacian spectrum.
\item 
\textbf{Gl2vec}~\cite{tu2019gl2vec}.
It constructs vectors for feature representations by comparing static and temporal network graphlet distributions to random graphs generated from different null models.
\item 
\textbf{Diffpool}~\cite{ying2018hierarchical}.
It is a differentiable graph pooling module that can generate hierarchical representations of graphs by learning a differentiable soft cluster assignment for vertices and maps them to a coarsened graph layer by layer.
This strategy is recently proposed and achieves the state-of-the-art effect in graph classification.
\end{itemize}
For all graph kernel and embedding methods, we implement graph classification by using the following machine learning classifiers: SVM based on radial basis kernel (SVM), Logistic regression classifier (Log), $k$-nearest neighbors classifier (KNN) and random forest classifier (RF). 
In total, there are $4\times 4 + 1=17$ available combinations of graph classification.

\subsection{Experiment Setup}
Each dataset is split into training, validation and testing sets with a proportion of 7:1:2. 
In this work, the validation set $D_{\textit {val}}$ takes effect in two parts: 
1) finetune the hyper-parameters of classifiers in combination with grid search;
2) compute the label reliability threshold and confusion matrix of classifiers during model evolution.
We repeat 5-fold cross validation for 10 times and report the average accuracy across all trials.

For all kernel and embedding methods, we set the embedding dimension to 128. 
We then set the budget of edge modification $\beta$ as 0.15.
During motif-random and motif-similarity mapping, we fix $l=2$, which means that we only consider open-triad motif.
Furthermore, during model evolution, we set the number of iterations $T$ to 5.
\subsection{Evaluation}\label{evaluation}
We evaluate the benefit of the proposed graph augmentation methods and {\em{M-Evolve}} framework, answering the following research questions:
\begin{itemize}
	\item      
	\textbf{RQ1}: Can {\em{M-Evolve}} improve the performance of graph classification when being combined with existing graph classification models?
	\item  
	\textbf{RQ2}: What are the roles that similarity and motif mechanisms play in improving the effect of graph augmentation?
	\item
	\textbf{RQ3}: Is the data filtration necessary and how does it help {\em{M-Evolve}} achieve performance improvement in graph classification?
	\item
	\textbf{RQ4}: How does {\em{M-Evolve}} achieve interpretable enhancement of graph classification?
\end{itemize}
We combine the proposed model evolution framework with all graph classification models to show a crosswise comparison.
Specifically, the four graph augmentation methods are combined with 17 graph classification combinations, totaling 68 available experimental combinations.
For instance, a practical combination could be \textit{\{vertex-similarity mapping + Graph2vec + KNN\}} or \textit{\{random mapping + Diffpool\}}.

\subsubsection{Enhancement for Graph Classification}
TABLE~\ref{tb:result} reports the results of performance comparison between the evolutive models and the original models, from which one can observe that
there is a significant boost in classification performance across all six datasets.
Overall, these models combined with the proposed {\em{M-Evolve}} framework obtain higher average classification accuracy in most cases and the {\em{M-Evolve}} achieves a 96.81\% success rate on the enhancement of graph classification
\footnote{The success rate refers to the percentage of evolutive models with accuracy higher than that of the corresponding original models in Table 3. 
The actual calculation formula is 395$\div$408=96.81\%.}.
These phenomena provide a positive answer to \textbf{RQ1}, indicating that the {\em{M-Evolve}} significantly improve the performance of the 17 graph classification combinations. 
We speculate that the original models trained with limited training data are over-fitting, and on the contrary, {\em{M-Evolve}} enriches the scale of training data via graph augmentation and optimizes graph classifiers via iterative retraining, which can improve the generalization and avoid over-fitting to a certain extent.

Now, we define the relative improvement rate (RIMP) in accuracy as follows:
\begin{equation}
	\textit{RIMP} = \frac{{\textit{Acc}}_{\textit{en}} - {\textit{Acc}}_{\textit{ori}}}{{\textit{Acc}}_{\textit{ori}}},
\end{equation}
where ${\textit{Acc}}_{\textit{en}}$ and ${\textit{Acc}}_{\textit{ori}}$ refer to the accuracy of the evolutive and original models, respectively.
Comparing all the graph augmentation methods, we count the numbers of experiments where they obtain the highest \textit{RIMP}, which were 1, 32, 24, 57 for {\em{random}}, {\em{vertex-similarity}}, {\em{motif-random}} and {\em{motif-similarity}}, respectively.
In TABLE~\ref{tb:result}, the far-right column gives the average relative improvement rate (Avg \textit{RIMP}) in accuracy, from which one can see that the {\em{M-Evolve}} combined with similarity-based mappings obtain the best results overall. In particular, {\em{motif-similarity mapping}}, which combines similarity and motif mechanisms, outperforms the others in half cases. These results indicate that both similarity and motif mechanisms play positive roles in enhancing graph classification, answering \textbf{RQ2}. 
As a reasonable explanation, similarity mechanism tends to link vertices with higher similarity and is capable of optimizing topological structure legitimately, which is similar to the finding in~\cite{zheng2020geometric}. 
And the motif mechanism achieves edge modification via local edge swapping, which has subtle effect on both the degree distribution and the clustering coefficient of the graph.

\begin{figure}[htb]
	\centering
	\includegraphics[width=\linewidth]{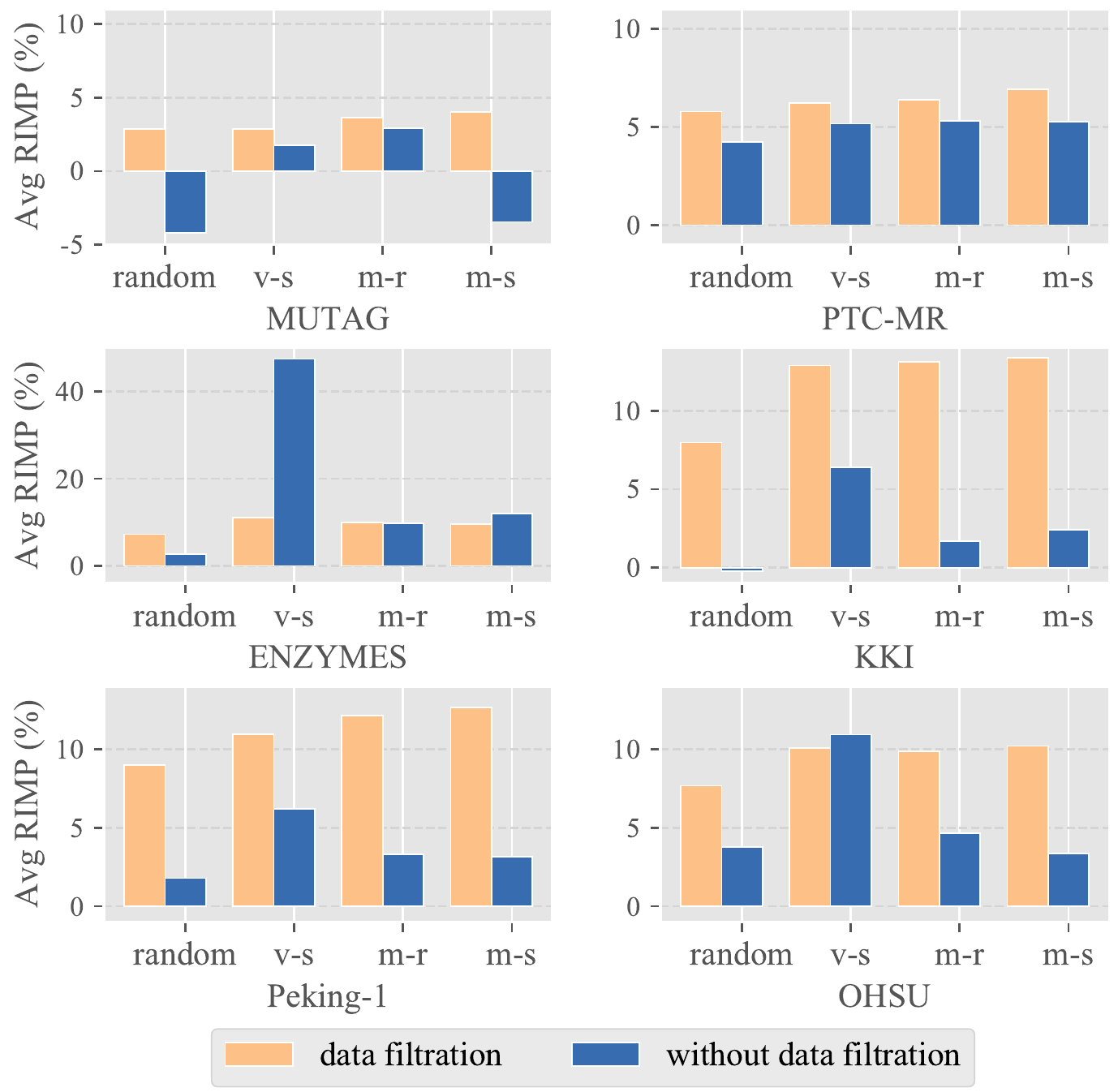}
	\caption{The impact of data filtration on classification performance in term of Avg IMP.}
	\label{fig:data-filter-bar}
	\vspace{-12pt}
\end{figure}
\begin{figure*}[htb]	
	\centering
	\includegraphics[width=\textwidth]{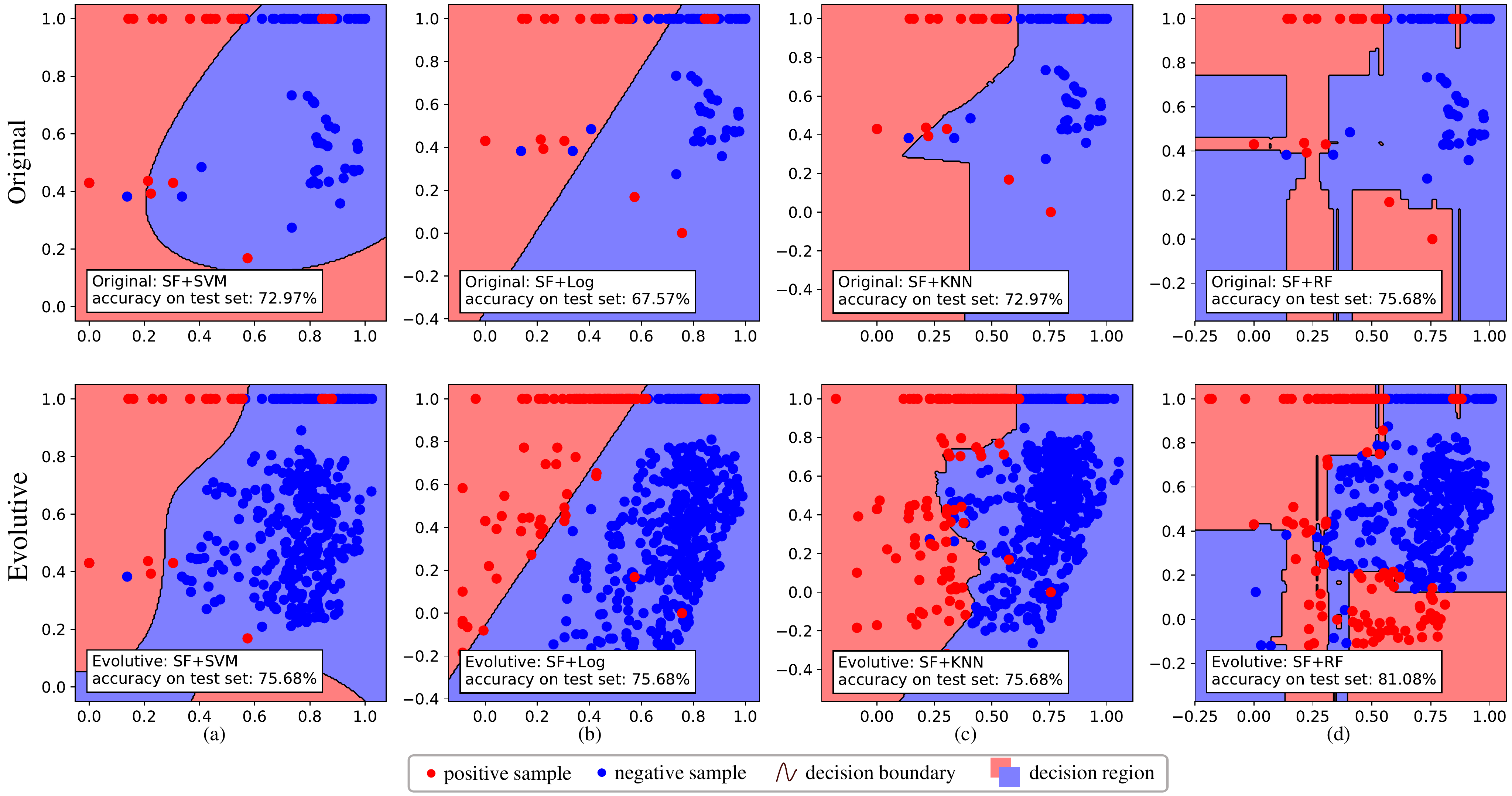}
	\caption{Visualization of training data distribution and decision boundaries of different graph classifiers on MUTAG dataset. 
			 Points with different colors represent training data with different labels and regions with different colors belong to different classes.}
	\label{fig:diff-clf-dr}
\end{figure*}
\begin{figure*}[!ht]	
	\centering
	\includegraphics[width=\textwidth]{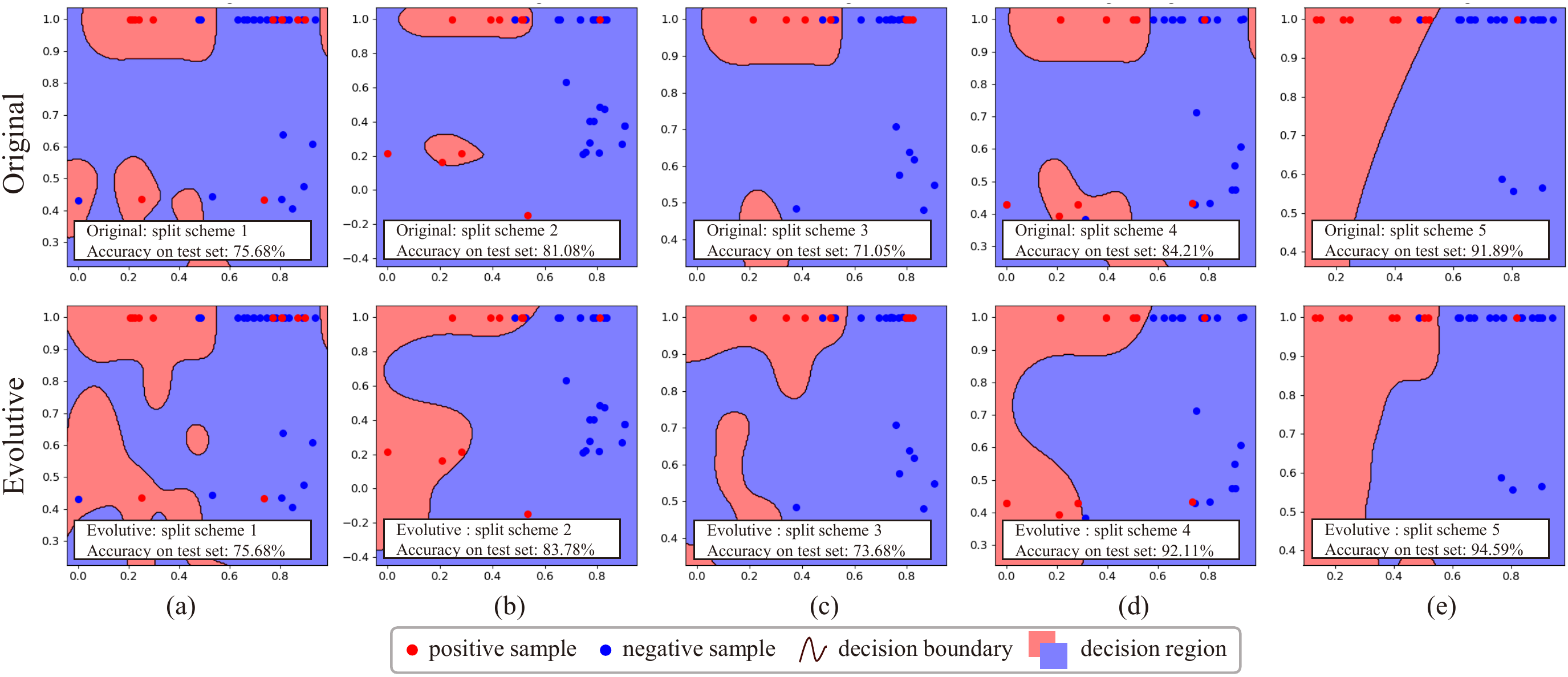}
	\caption{Visualization of decision boundaries of graph classification combination (SF + SVM) on MUTAG dataset. Points with different colors represent testing data with different labels.}
	\label{fig:sfsvm-clf-dr}
\end{figure*}
\begin{figure*}[htb]
	\centering
	\includegraphics[width=\textwidth]{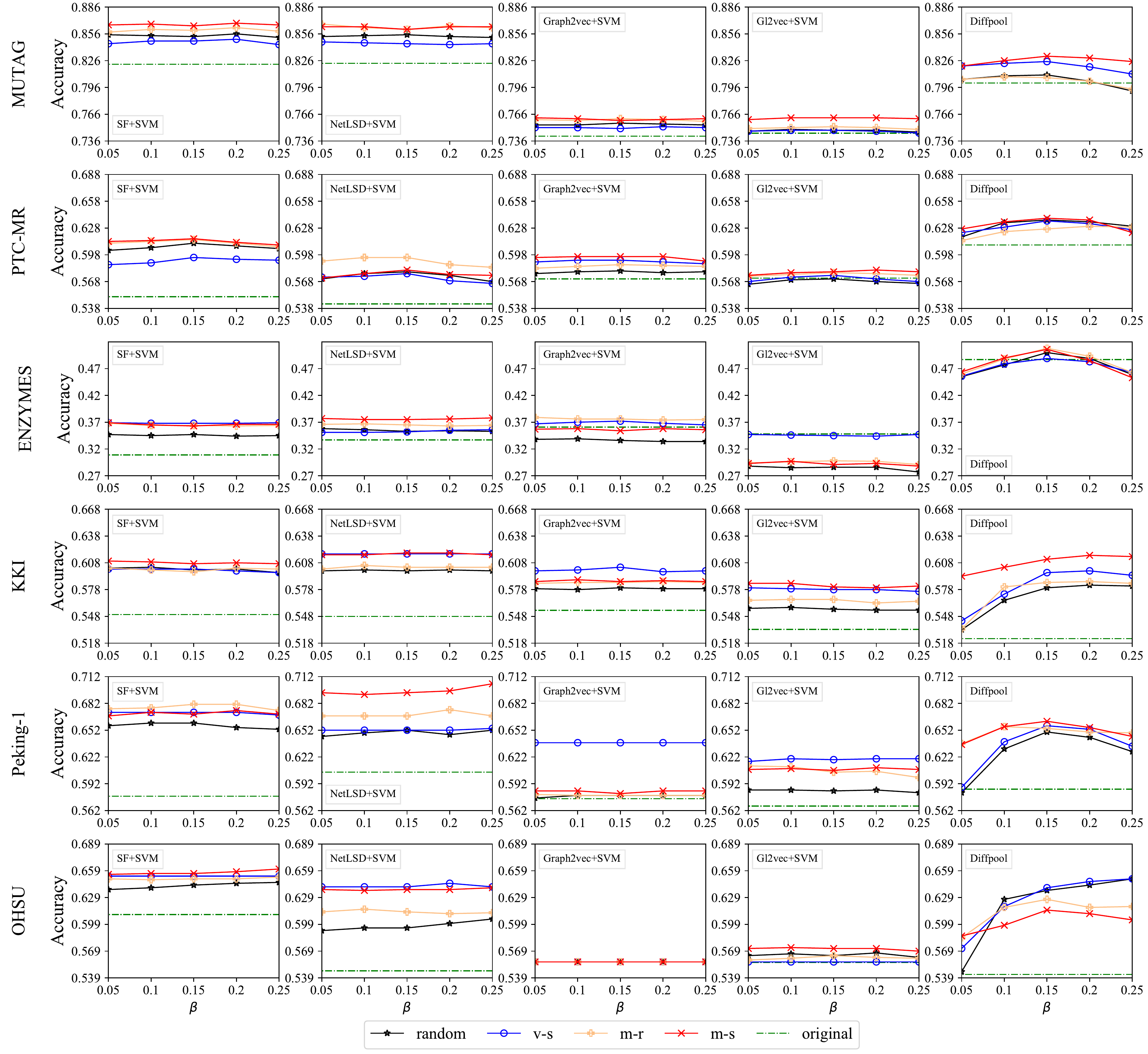}
	\caption{Parameter sensitivity evaluation of the {\em{M-Evolve}} framework on graph classification.}
	\label{fig:diff-ratio}
\end{figure*}
\subsubsection{Impact of Data Filtration in Model Evolution} \label{sec:impact-of-data-filtration}
Thanks to the outstanding performance of {\em{M-Evolve}}, we further investigate the impact of the data filtration on the enhancement of graph classification.
Specifically, we conduct contrastive experiments in which the data filtration operation is removed from {\em{M-Evolve}} and the performance differences between the two cases with and without data filtration are shown in Fig.~\ref{fig:data-filter-bar}.
From the comparison results, we observe that there is a more significant improvement in classification performance when the {\em{M-Evolve}} framework is combined with data filtration in most cases, positively answering \textbf{RQ3}.

Notably, without data filtration, the results of {\em{M-Evolve}} with different mappings vary considerably in overall quality, and this mechanism may even have negative effect in certain cases. On the contrary, with data filtration, {\em{M-Evolve}} has more significant and consistent effects on enhancing graph classification among different mappings.
A reasonable explanation for this effect is that data filtration is capable of retaining examples that are conducive to model's decision, so that these augmented sets obtained via various mappings tend to have similar feature distributions. 
As a result, data filtration narrows the quality gap between the data generated by {\em{random mapping}} and those by the other three mappings, achieving the consistency of performance. 
As an exception, the {\em{vertex-similarity mapping}} (v-s) shows particularly excellent performance in the case without data filtration in ENZYMES.
One possible explanation is that accepting more augmented examples may be more favorable for optimizing relatively complex multiclass classification.
\subsubsection{Explanatory Visualization of Data and Models}
Next, we apply visualization techniques to investigate how graph augmentation enriches the data distribution and how the {\em{M-Evolve}} framework optimizes the performances of different models.
Since higher-dimensional data are difficult to visualize, we set the embedding dimension of both graph kernel and embedding models to 2.

Firstly, we compare the training data distribution before and after model evolution, as visualized in Fig.~\ref{fig:diff-clf-dr}, to demonstrate the effectiveness of the proposed graph augmentation.
Specifically, the top and bottom rows show the decision regions of the original and evolutive models respectively, and the points with different colors represent training data with different labels, (a)$\sim$(d) are based on the same data split and graph augmentation, but different graph classification combinations (\textit{vertex-similarity mapping + SF + \{SVM, Log, KNN, RF\}}). Obviously, there is a significant boost in the scale of training data and the distribution boundaries of data with different labels, indicating that graph augmentation effectively enriches the training data and the new data distribution is more conducive to the training of classifiers. 

Furthermore, we visualize the decision boundaries in Fig.~\ref{fig:sfsvm-clf-dr}, to clearly highlight the difference between the original and evolutive models.
Specifically, (a)$\sim$(e) are based on the same combination (\textit{vertex-similarity mapping + SF + SVM}), but different data split schemes, which refer to testing using different folds of dataset.
As one can see, the decision regions of the non-dominant class are fragmented and scattered in the original models. 
During model evolution, scattered regions tend to merge, and the original decision boundaries are optimized to smoother ones.
These phenomena answer \textbf{RQ4}.

In summary, graph augmentation can efficiently increase the data scale, indicating its ability in enriching data distribution. 
And the entire {\em{M-Evolve}} framework is capable of optimizing the decision boundaries of the classifiers and ultimately improving their generalization performances.
\begin{figure}[htb]	
	\centering
	\includegraphics[width=\linewidth]{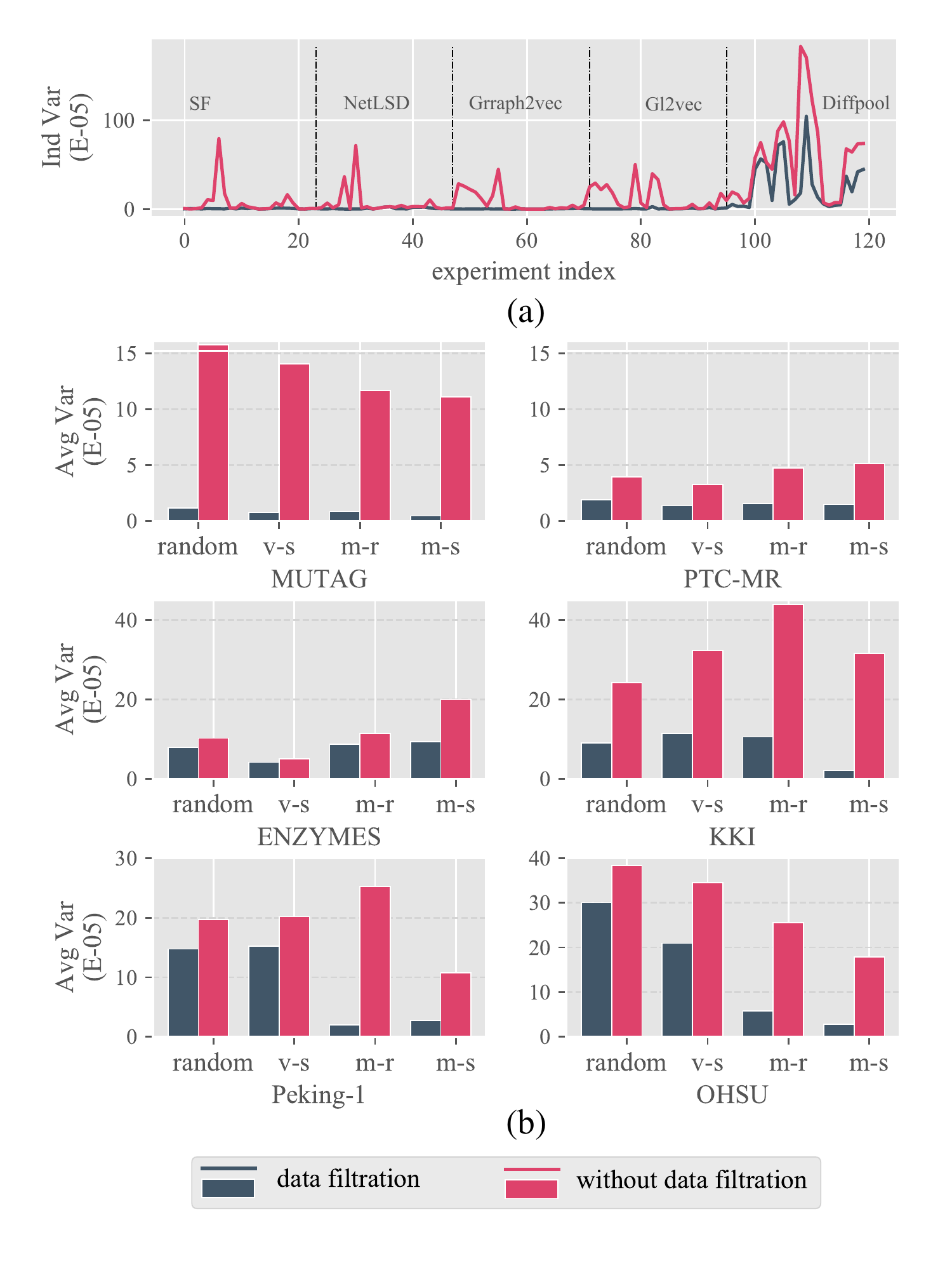}
	\caption{The impact of data filtration on the parameter sensitivity.}
	\label{fig:var}
	\vspace{-5pt} 
\end{figure}

\subsubsection{Parameter Sensitivity}
In this subsection, we further analyze the impact of key parameters on the performance of the {\em{M-Evolve}} framework. 
Specifically, we vary the budget of edge modification $\beta$ in $\{0.05,0.1,0.15,0.2,0.25\}$.
We present the evaluation results of graph classification based on these combinations (\textit{\{SF, NetLSD, Graph2vec, Gl2vec\} + SVM} \& \textit{Diffpool}) in Fig.~\ref{fig:diff-ratio}, involving the graph kernel and embedding methods with SVM and the GNN-based Diffpool method.
From the results, one can see that the {\em{M-Evolve}} framework is not strictly sensitive to different parameter settings when it comes to graph kernel and embedding models.
On the other hand, when it comes to the Diffpool model, there are consistent tendencies in the sensitivity curves among all datasets, indicating that too large or too small perturbations are not conducive to graph augmentation.

Furthermore, we supplement the contrastive experiments of parameter sensitivity without data filtration, as shown in Fig.~\ref{fig:var}, to verify our conjecture that data filtration can actually help {\em{M-Evolve}} reduce parameter sensitivity.
Specifically, we use the variance of the five data points on each curve to measure the corresponding parameter sensitivity. 
Fig~\ref{fig:var} (a) shows the individual variance (Ind Var) of all the experimental combinations involved in Fig.~\ref{fig:diff-ratio}, and (b) presents the average variance for each dataset.
From the comparison results, one can observe that these results with data filtration have less fluctuation and better stability under different parameter settings, which
provide positive support for our conjecture that data filtration can actually improve the robustness of the {\em{M-Evolve}} against parameter variations.
As for the more obvious fluctuations on the curves of Diffpool, we speculate that end-to-end deep learning models are more capable of capturing slight changes in data features  when compared to machine learning models like SVM.

In summary, data filtration not only narrows the performance gap among different graph augmentation methods, but also reduces the sensitivity of {\em{M-Evolve}} to different parameter settings, 
implying that the {\em{M-Evolve}} framework is robust to parameter settings to a certain extent, and thus could be more easily applied in reality.

\section{Conclusion} \label{sec:Conclusion}
In this paper, we introduce data augmentation for graph classification and present four heuristic algorithms to generate weakly labeled data for small-scale benchmark datasets via a heuristic transformation of the graph structure. 
Furthermore, we propose a generic model evolution framework named {\em{M-Evolve}}, which combines graph augmentation, data filtration and model retraining to optimize pre-trained graph classifiers. 
Experiments conducted on six benchmark datasets demonstrate that our proposed framework performs surprisingly well and helps existing graph classification models alleviate over-fitting when training on small-scale benchmark datasets 
and achieve significant improvement of classification performance.
For future work, we will design effective graph augmentation methods on large scale graphs and extend the current framework to work on real-world datasets like social networks and transaction networks.



\ifCLASSOPTIONcompsoc
  \section*{Acknowledgments}
\else
  \section*{Acknowledgment}
\fi
The authors would like to thank all the members in the IVSN Research Group, Zhejiang University of Technology for the valuable discussions about the ideas and technical details presented in this paper. This work was partially supported by the National Natural Science Foundation of China under Grant 61973273, by the Zhejiang Provincial Natural Science Foundation of China under Grant LR19F030001, and by the Hong Kong Research Grants Council under the GRF Grant CityU11200317.




\bibliographystyle{IEEEtran}
\bibliography{MyBibliography}




%
\vspace{-35pt} 
\begin{IEEEbiography}[{\includegraphics[width=1in,height=1.25in,clip,keepaspectratio]{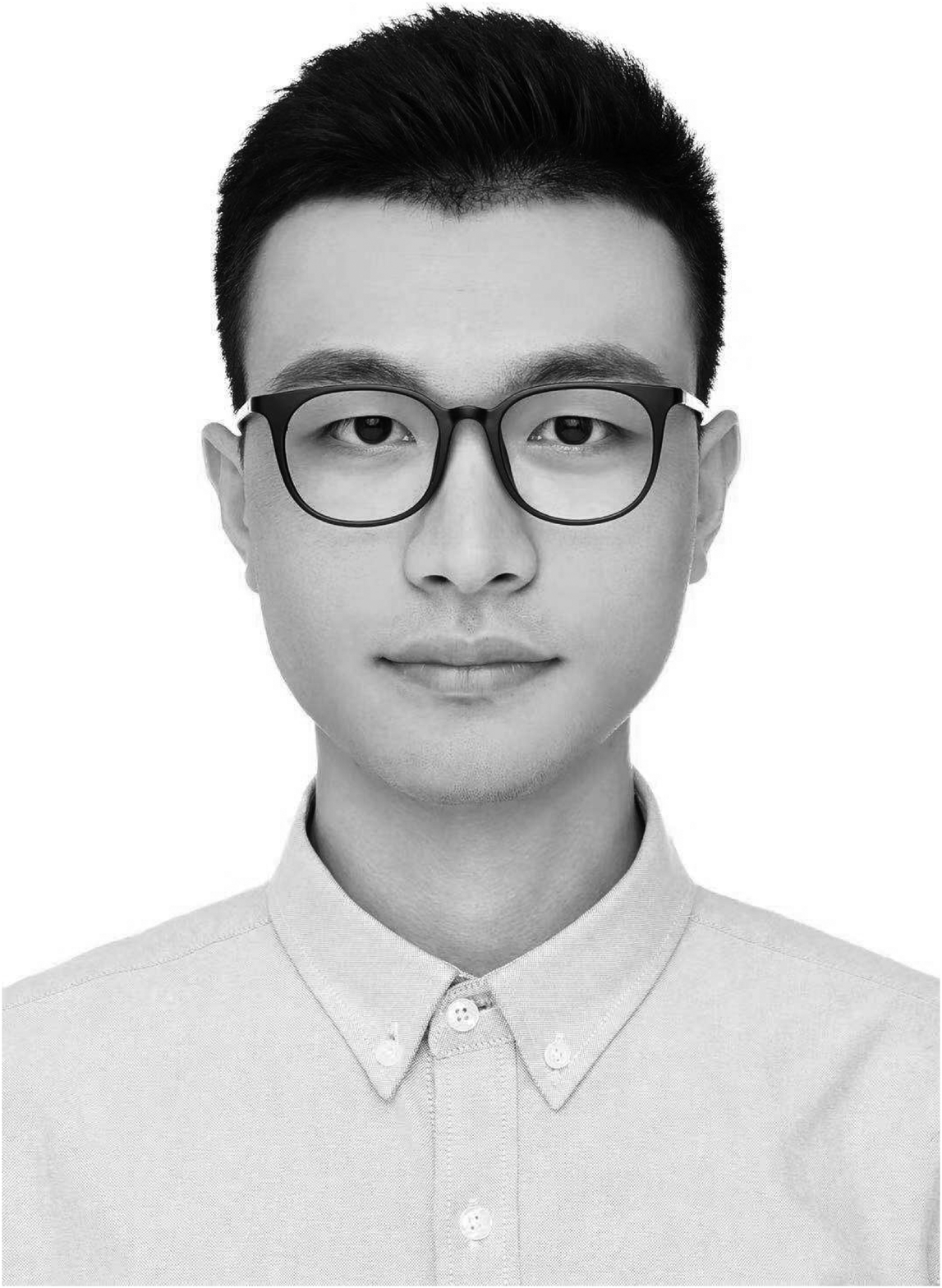}}]{Jiajun Zhou}
received the BS degree in automation from the Zhejiang University of Technology, Hangzhou, China, in 2018, where he is currently pursuing the MS degree in control theory and engineering with the College of Information and Engineering.
His current research interests include graph mining and deep learning, especially for network security and knowledge graph reasoning.
\end{IEEEbiography}
\vspace{-30pt} 
\begin{IEEEbiography}[{\includegraphics[width=1in,height=1.25in,clip,keepaspectratio]{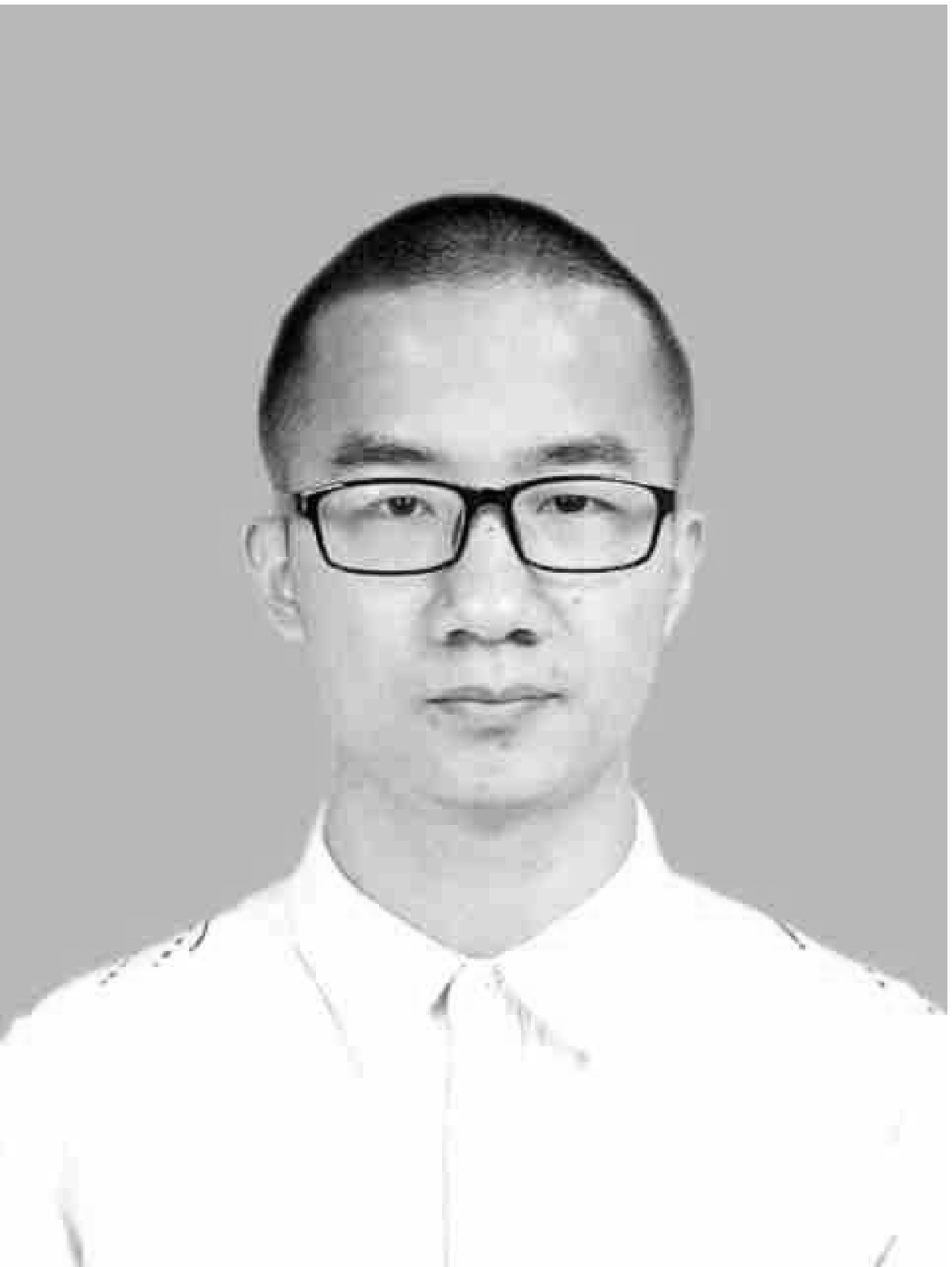}}]{Jie Shen} received the BS degree in automation from Zhejiang University of Science and Technology, Hangzhou, China, in 2019. He is currently pursuing the MS degree at the College of Information Engineering, Zhejiang University of Technology, Hangzhou, China. His current research interests include graph mining and network security.
\end{IEEEbiography}
\vspace{-30pt} 
\begin{IEEEbiography}[{\includegraphics[width=1in,height=1.25in,clip,keepaspectratio]{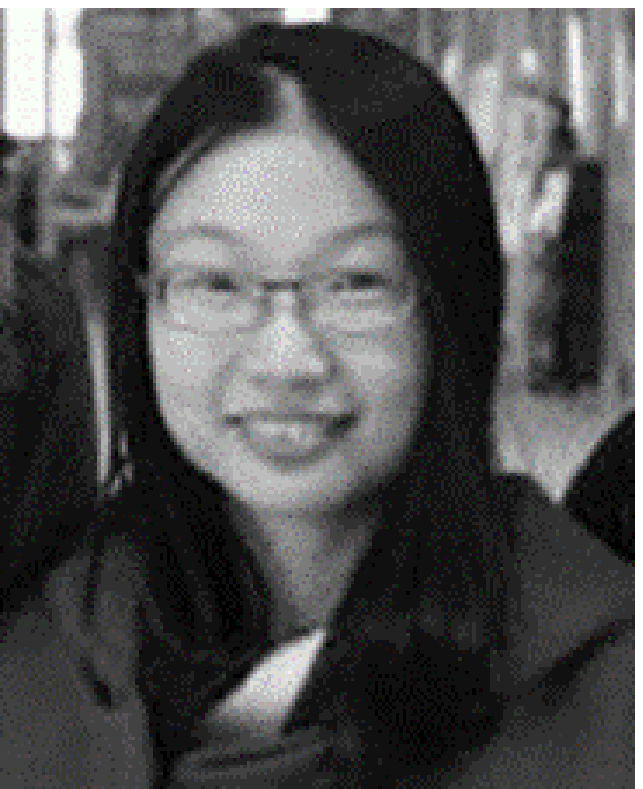}}]{Shanqing Yu} received the MS and PhD degrees from the Graduate School of Information, Production and Systems, and the School of Computer Engineering, Waseda University, Japan, in 2008 and 2011, respectively. She is currently a lecturer at the College of Information Engineering, Zhejiang University of Technology. Her research interests cover intelligent computation, data mining and intelligent transport systems.
\end{IEEEbiography}
\vspace{-30pt} 
\begin{IEEEbiography}[{\includegraphics[width=1in,height=1.25in,clip,keepaspectratio]{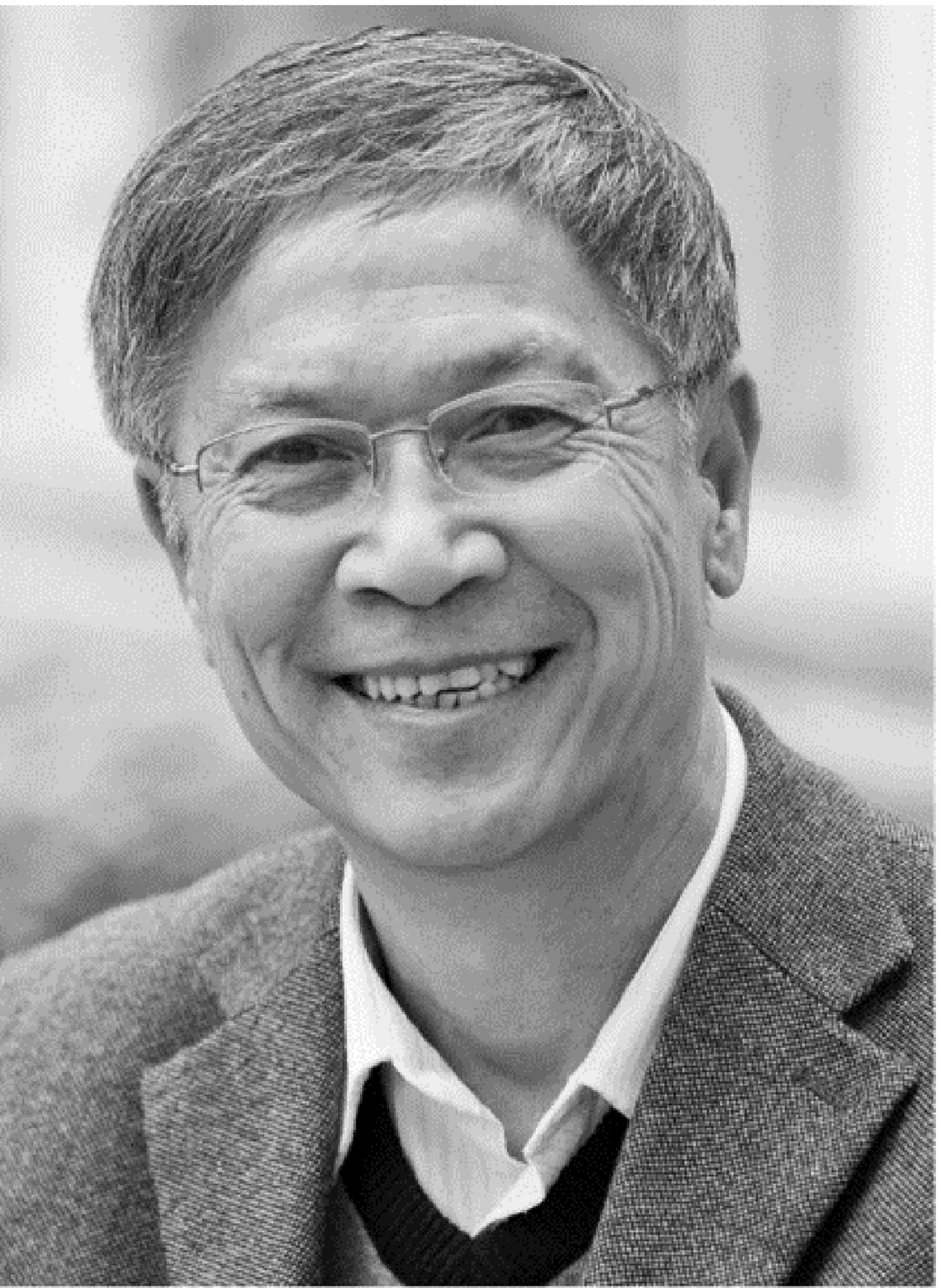}}]{Guanrong Chen}(M'89-SM'92-F'97) 
received the MSc degree in computer science from Sun Yat-sen University, Guangzhou, China in 1981, and the PhD degree in applied mathematics from Texas A\&M University, College Station, Texas, in 1987. 
He has been a chair professor and the founding director of the Centre for Chaos and Complex Networks at the City University of Hong Kong since 2000. 
Prior to that, he was a tenured full professor with the University of Houston, Texas. He was awarded the 2011 Euler Gold Medal, Russia, and conferred a Honorary Doctorate by the Saint Petersburg State University, Russia in 2011 and by the University of Le Havre, Normandy, France in 2014. 
He is a member of the Academy of Europe and a fellow of The World Academy of Sciences, and is a Highly Cited Researcher in Engineering as well as in Mathematics according to Thomson Reuters.
\end{IEEEbiography}
\vspace{-30pt} 
\begin{IEEEbiography}[{\includegraphics[width=1in,height=1.25in,clip,keepaspectratio]{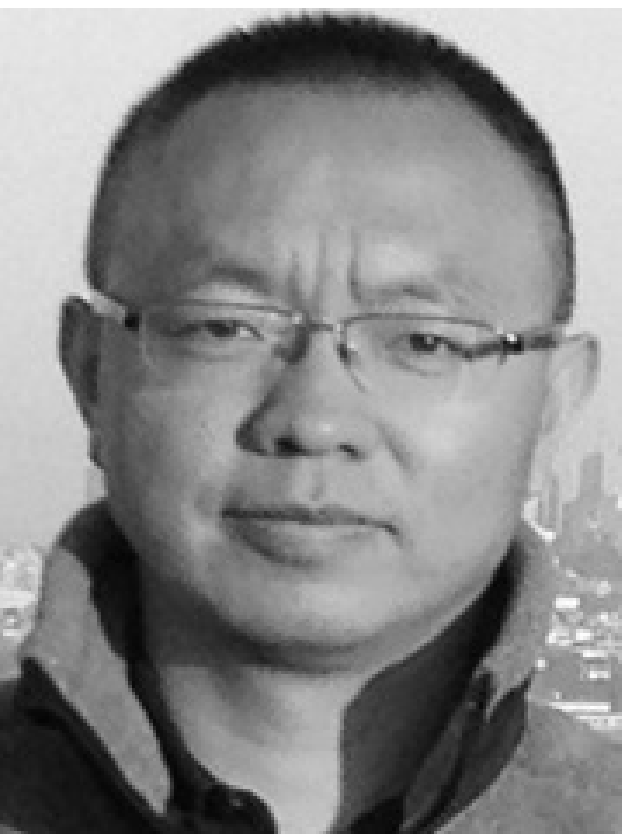}}]{Qi Xuan}(M'18) received the BS and PhD degrees in control theory and engineering from Zhejiang University, Hangzhou, China, in 2003 and 2008, respectively. He was a Post-Doctoral Researcher with the Department of Information Science and Electronic Engineering, Zhejiang University, from 2008 to 2010, respectively, and a Research Assistant with the Department of Electronic Engineering, City University of Hong Kong, Hong Kong, in 2010 and 2017. From 2012 to 2014, he was a Post-Doctoral Fellow with the Department of Computer Science, University of California at Davis, CA, USA. He is a member of the IEEE and is currently a Professor with the Institute of Cyberspace Security, College of Information Engineering, Zhejiang University of Technology, Hangzhou, China. His current research interests include network science, graph data mining, cyberspace security, machine learning, and computer vision.
\end{IEEEbiography}







\end{document}